\documentclass[letterpaper]{article} 
\usepackage{aaai2027}  
\usepackage[hyphens]{url}  
\usepackage{graphicx} 
\usepackage{natbib}  
\usepackage{caption} 
\usepackage{algorithm}
\usepackage{algorithmic}

\usepackage{newfloat}
\usepackage{listings}
\DeclareCaptionStyle{ruled}{labelfont=normalfont,labelsep=colon,strut=off} 
\floatstyle{ruled}
\newfloat{listing}{tb}{lst}{}
\floatname{listing}{Listing}

\usepackage{booktabs}

\usepackage{placeins}
\usepackage{multirow}
\usepackage{arydshln}
\usepackage{tikz}
\usepackage{makecell}
\usepackage{pifont}
\usepackage{threeparttable}
\usepackage[table,xcdraw]{xcolor}
\usepackage{subcaption}
\usepackage{amssymb}
\usepackage{amsmath}
\usepackage{array} 
\usepackage{caption}
\title{Risk-Aware Occupancy for Safety-Oriented End-to-End Autonomous Driving}

\author{
Jiaxing Chen\textsuperscript{1},
Hengduo Zou\textsuperscript{1},
Yiren Zhao\textsuperscript{2},
Bolin Gao\textsuperscript{1}\thanks{$\dagger$ Corresponding author: Bolin Gao}
}
\affiliations{
\textsuperscript{1}School of Vehicle and Mobility, Tsinghua University\\
\textsuperscript{2}The Hong Kong University of Science and Technology (Guangzhou)
}
\nocopyright

\begin{document}

\maketitle
\begin{abstract}

Conventional end-to-end driving systems model the environment with sparse objects and lane elements. While efficient, this paradigm discards planning-critical information in crowded and occluded scenarios, particularly for unstructured obstacles, ambiguous free space, and complex interactions. We propose risk-aware occupancy, a dense BEV representation that explicitly fuses geometric occupancy, map-derived traffic constraints, and future dynamic-agent occupancy as complementary risk signals. Built upon this representation, we develop ROIDrive, an instance-centric end-to-end framework with a dedicated risk-aware occupancy branch. The predicted occupancy is tokenized via sliding-window sampling and injected into planning queries via cross-attention, while temporal query consistency mitigates unreliable flickering queries. We also contribute RiskOcc4D-nuScenes, a benchmark derived from nuScenes and Occ3D-nuScenes with four automated annotation pipelines for multi-dimensional risk supervision. Experiments on representative occupancy architectures verify the learnability and transferability of our representation. Integrated with GenAD, it reduces collision rates by 35.0\% (UniAD metric) and 52.9\% (ST-P3 metric), confirming the efficacy of the proposed representation modality.

\end{abstract}
\section{Introduction}

Safety remains a central barrier to deploying SAE Level~2 and Level~4 autonomous driving. Although end-to-end methods have advanced rapidly, sparse agent and map representations can miss unstructured obstacles, occluded regions, and uncertain free space. Dense spatial fields instead allow planners to evaluate trajectory feasibility and nearby conflicts directly. We therefore propose risk-aware occupancy to complement existing E2E frameworks with explicit safety cues.

\begin{figure*}[t!]
    \centering
    \includegraphics[width=0.99\linewidth,
        trim=1 1 1 1,
        clip]{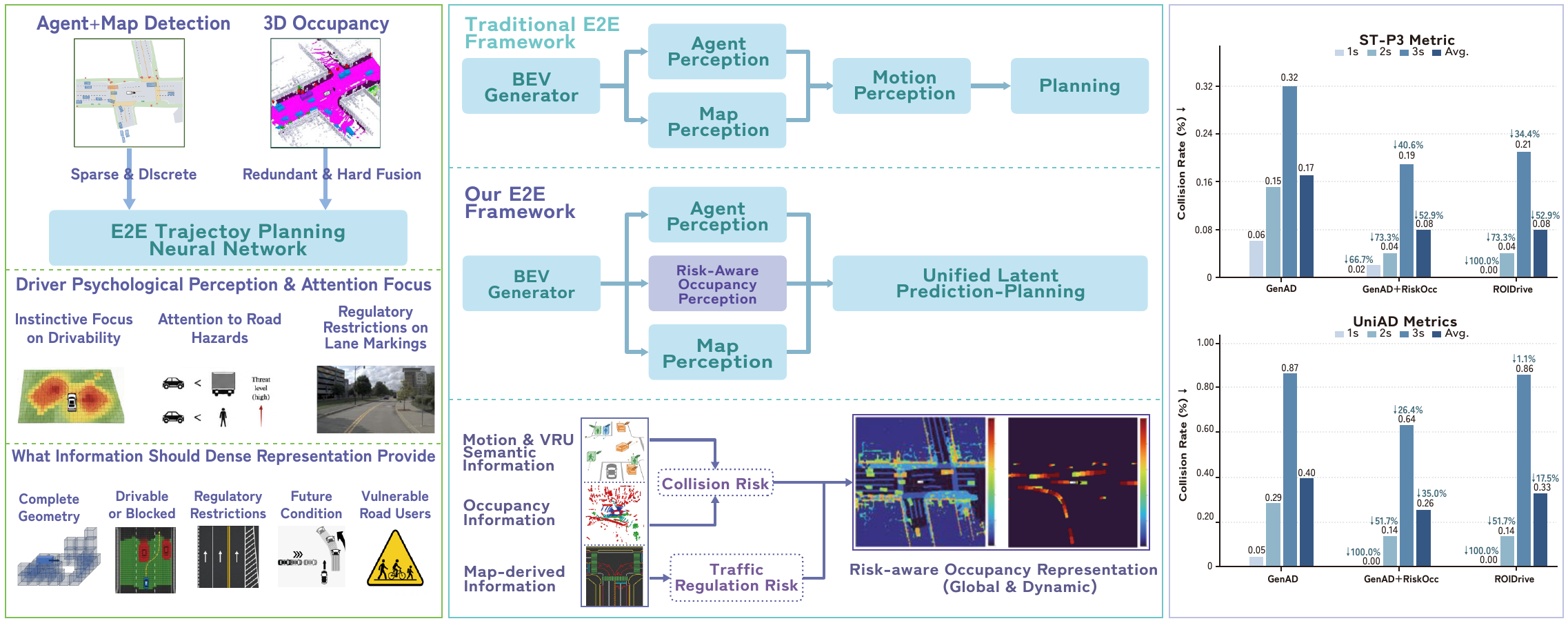}
    \caption{Motivation and design of risk-aware occupancy.
    Left: Limitations of sparse agent/map detection and naive 3D occupancy, and human-inspired requirements for dense representations covering geometry, drivability, traffic rules and vulnerable road users.
    Middle: Comparison between conventional decoupled E2E pipelines and our unified prediction-planning framework, which constructs global and dynamic risk-aware occupancy maps by encoding collision and traffic-regulation risks.
    Right: Collision-rate benchmarks on UniAD and ST-P3 metrics, verifying the safety improvement of the proposed risk-aware occupancy representation.}
    \label{fig:risk_occ_overview}
\end{figure*}

Safety-critical driving scenarios motivate two core requirements for dense representations: continuously modeling arbitrarily-shaped drivable regions and capturing evolving spatial conflicts (Figure~\ref{fig:risk_occ_overview}). This aligns with human driving intuition, where safe passage is assessed before reasoning about fine-grained semantic categories~\cite{influence_line,field_drivers}. Physical traversability is encoded via occupied height, legal motion space via map priors, and imminent hazards via future agent motion, while semantic cues further refine traffic-rule compliance and hazard assessment~\cite{stress,impactdrivers}. Risk-aware Occupancy unifies these multi-source cues into a planner-accessible, interpretable representation.

Building upon prior risk occupancy modeling~\cite{riskoccupancy}, each BEV cell is assigned a calibrated risk score aggregating geometry, semantic priors, map constraints, and future motion. Projecting 3D information into BEV discards redundant vertical details and fits the feature layout of modern end-to-end networks. As such, Risk-aware Occupancy serves both as a perception supervision target and an intermediate planning representation, aligning perception objectives with downstream decision-making and enabling safety inspection, failure localization, and visual debugging.

This three-branch perception architecture is instantiated in ROIDrive, an instance-centric end-to-end driving framework. Beyond the conventional sparse object detection and lane-level map branches, a dedicated third branch predicts the dense Risk-aware Occupancy representation. Tokenized via sliding-window sampling, the dense occupancy tokens are fused with the original sparse instance and map tokens, and jointly injected into planning queries through cross-attention, enabling end-to-end joint optimization of all three perception streams with downstream instance-centric planning.

For empirical validation, we present \textit{RiskOcc4D-nuScenes}, a benchmark derived from Occ3D-nuScenes via four automated annotation pipelines that incorporate road and temporal cues under occlusion and point-cloud sparsity. Benchmark evaluations across multiple occupancy architectures characterize representation learnability and cross-model compatibility. Oracle-versus-predicted ablation studies decouple the intrinsic merit of the proposed representation from prediction errors, rigorously verifying its utility for downstream planning.

The main contributions of this work are summarized as follows:
\begin{enumerate}
    \item \textbf{Risk-aware occupancy representation.}
    We introduce a safety-oriented dense BEV representation that unifies geometry,
    safety-relevant semantics, map constraints, and future motion into an explicit
    planning-relevant risk-aware occupancy.
    
    \item \textbf{ROIDrive E2E framework.}
    We design ROIDrive, an E2E driving framework with an independent risk-aware occupancy branch seamlessly integrated at the planning stage, and validate it throughout the complete perception-to-planning pipeline.
    
    \item \textbf{RiskOcc4D-nuScenes dataset.}
    We construct the RiskOcc4D-nuScenes dataset using four automated annotation pipelines and instantiate risk-aware occupancy in the end-to-end framework to verify its effectiveness and quantify its performance improvements.
\end{enumerate}

\section{Related Work}
\label{sec:relatedworks}

\subsection{Occupancy Prediction Methods}

Recently, occupancy prediction has advanced rapidly along three axes. \textbf{Lightweight architectures}: FlashOcc~\cite{yu2023flashocc}, TPVFormer~\cite{huang2023tri}, SparseOcc~\cite{tang2024sparseocc}, FastOcc~\cite{hou2024fastocc}, and Octree-Occ~\cite{lu2023octreeocc} reduce cost via BEV extraction, tri-plane decomposition, sparse encoding, and octree reconstruction, respectively. \textbf{Performance optimization}: FB-Occ~\cite{li2023fb} fuses forward-backward projection~\cite{huang2021bevdet,li2022bevformer} with multi-task learning; F-OCC~\cite{zhao3D} extends DCNv4~\cite{xiong2024efficient} to 3D; OccTransformer~\cite{liu2024occtransformer}, OccFormer~\cite{zhang2023occformer}, VoxelFormer~\cite{li2023voxformer}, and MambaOcc~\cite{tian2024mambaocc} leverage advanced attention or state-space architectures. \textbf{Cross-domain integration}: NeRF-based methods~\cite{pan2023uniocc,zhang2023occnerf,huang2024selfocc,mildenhall2021nerf}, Gaussian splatting variants~\cite{huang2024gaussianformer,huang2024probabilistic,gan2024gaussianocc,kerbl20233d}, world models~\cite{zheng2025occworld}, and LLM-driven approaches~\cite{xu2025occ,wei2024occllama} further enrich occupancy representations from complementary perspectives.




 \begin{figure*}[t!]
  \centering
  \includegraphics[width=0.99\textwidth, trim=15 15 15 15, clip]{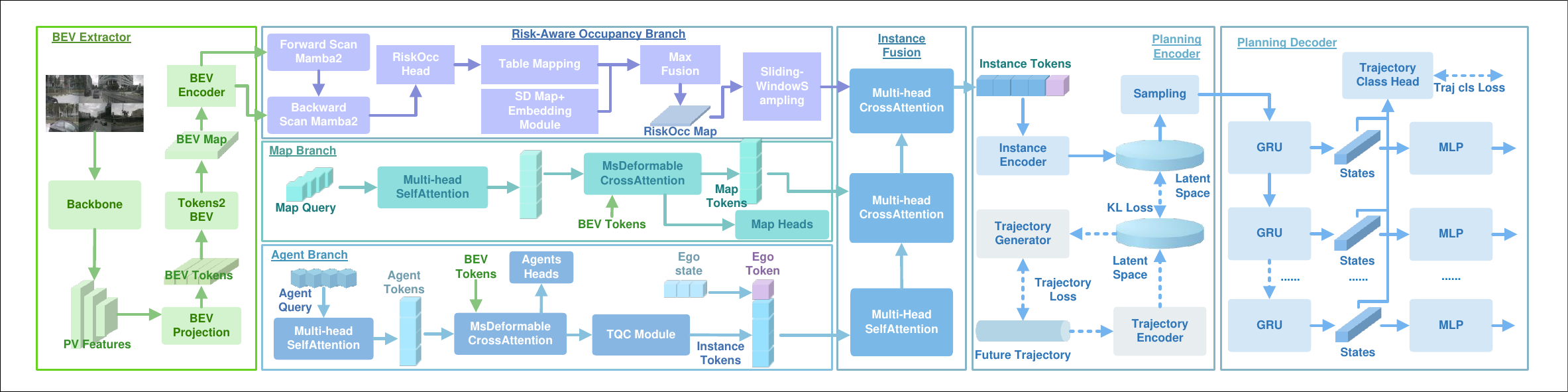}
  \caption{ROIDrive architecture. A BEVFormer backbone yields the BEV feature $B$; the risk-aware occupancy (RiskOcc) branch regresses a normalized risk-aware occupancy via a Bi-Mamba2 block and a class-to-risk lookup table (optionally fused with an SDMap+ prior), tokenizes it by sliding-window sampling, and injects it into the queries via cross-attention. Temporal Query Consistency (TQC) then gates the queries by cross-frame reliability before a VAE generative planner decodes multimodal trajectories.}
  \label{fig:roi}
\end{figure*}

\subsection{Auxiliary Perception Supervision in End-to-End Driving Models}
End-to-end driving models often regularize shared BEV features with auxiliary detection, segmentation, mapping, or occupancy tasks. VADv2~\cite{VADv2} and DiffusionDriveV2~\cite{DiffusionDriveV2} use BEV detection; HENet++~\cite{HENet++} adds map segmentation; OccVLA~\cite{OccVLA} and IR-WM~\cite{IRWM} use volumetric occupancy; and DrivoR~\cite{DrivoR} removes dense occupancy for efficiency. These objectives are not directly aligned with trajectory ranking, while full 4D occupancy remains costly. Risk-aware occupancy instead provides a compact, trajectory-aligned spatiotemporal risk signal.

\section{Risk-Aware Occupancy Perception}
\label{sec:paradigm}

Conventional 3D semantic occupancy builds high-entropy voxel
representations $H(\mathcal{V})$ with fine-grained categories;
though complete, they carry substantial task-irrelevant
redundancy $H_{\mathrm{red}}(\mathcal{V})$ that degrades
risk-information transfer for E2E driving.

Risk-aware occupancy prediction is formulated as
planning-oriented minimal sufficient statistic extraction:
\begin{equation}
  R^{*} = \arg\max_{R}\; I(R;\, T_{\mathrm{plan}}),
  \quad \text{s.t.}\;\; \dim(R) \ll \dim(\mathcal{V}),
  \label{eq1}
\end{equation}
where $I(R;\,T_{\mathrm{plan}})$ is the mutual information
between risk representation $R$ and the optimal trajectory
$T_{\mathrm{plan}}$.

Moreover, conventional 3D occupancy is overcomplete: its
semantic paradigm cannot uniformly quantify collision
threats and regulatory constraints
(\emph{e.g.}, lane markings, stop lines). Risk-aware
occupancy encodes both risk types and resolves conflicts
via maximum aggregation over collision and regulatory
terms. To realize this formulation, dual-axis reduction
is performed:
\begin{itemize}
  \item \textbf{Spatial:} Only the critical height
  interval $H_{\mathrm{crit}}\in[0\;\mathrm{m},\,
  2.2\;\mathrm{m}]$ (the vehicle-relevant collision
  zone) is retained, discarding non-threatening
  airspace.
  \item \textbf{Semantic:} Fine-grained labels are
  replaced by object height as the primary risk
  quantifier; vulnerable road users (VRUs) retain
  semantic priors with elevated risk weights.
\end{itemize}

A hierarchical risk quantification framework is
constructed over the Occ3D-nuScenes perception domain
$\Omega=\{(x,y)\mid x,y\in[-40\;\mathrm{m},\,
40\;\mathrm{m}]\}$. Critical heights are nonlinearly
mapped to a normalized risk score
$Z_{\mathrm{norm}}\in[0,1]$. To separate risk modes, static
global obstacle risk is bounded to
$V_{\mathrm{risk}}^{\mathrm{global}}\in[0.2,\,0.7]$ and
moving objects to a dedicated branch with
$V_{\mathrm{risk}}^{\mathrm{dyn}}\in[0.5,\,1.0]$. Their
deliberate overlap enforces a strict priority
hierarchy---dynamic collision $>$ static obstacle $>$
regulatory risk---so dynamic threats receive planning
priority while gradient conflict in multi-task optimization
is mitigated.

\subsection{ROIDrive Framework}

\noindent\textbf{Design rationale.} As illustrated in Figure~\ref{fig:roi}, ROIDrive builds on the query-based generative E2E paradigm \cite{genad,jiang2023vad}, where a BEVFormer backbone yields a BEV feature $B\in\mathbb{R}^{C\times H\times W}$, a small fixed set of agent and map queries abstracts the scene, and a conditional VAE planner decodes multimodal trajectories. This compression causes two structural weaknesses: \emph{representation sparsity}, as the dense, temporally evolving risk structure of the BEV plane is left implicit, and \emph{token sparsity and unreliability}, as missed detections erase risk while false-positive and flickering queries inject spurious motion intent. ROIDrive thus pursues two goals: (1) to introduce a perception modality that is independently predicted and then fused into the E2E pipeline, and (2) to densify the representation and calibrate token reliability. Both are met by two modules at the motion-prediction stage: a risk-aware occupancy (RiskOcc) branch and a Temporal Query Consistency (TQC) module.

\noindent\textbf{Risk-aware occupancy branch.} The BEV feature is channel-reduced, modeled by a Bi-Mamba2 state-space block that captures long-range dependencies at $\mathcal{O}(N)$ cost, and upsampled. 
A lookup table converts the predicted risk classes into a normalized risk-aware occupancy $R\in[0,0.7]$; after optional fusion with an SDMap+ prior, $R$ is tokenized by sliding-window sampling and injected into the queries via cross-attention. The field complements the sparse queries: the queries encode object identity and location, whereas the field encodes current and near-future risk at each location. Because the branch predicts the dense field directly from $B$ rather than from detected objects, it also provides a fallback for missed hazards and anticipates risk through long-range modeling and future supervision.

\begin{figure*}[t]
    \centering
    \includegraphics[width=0.99\linewidth]{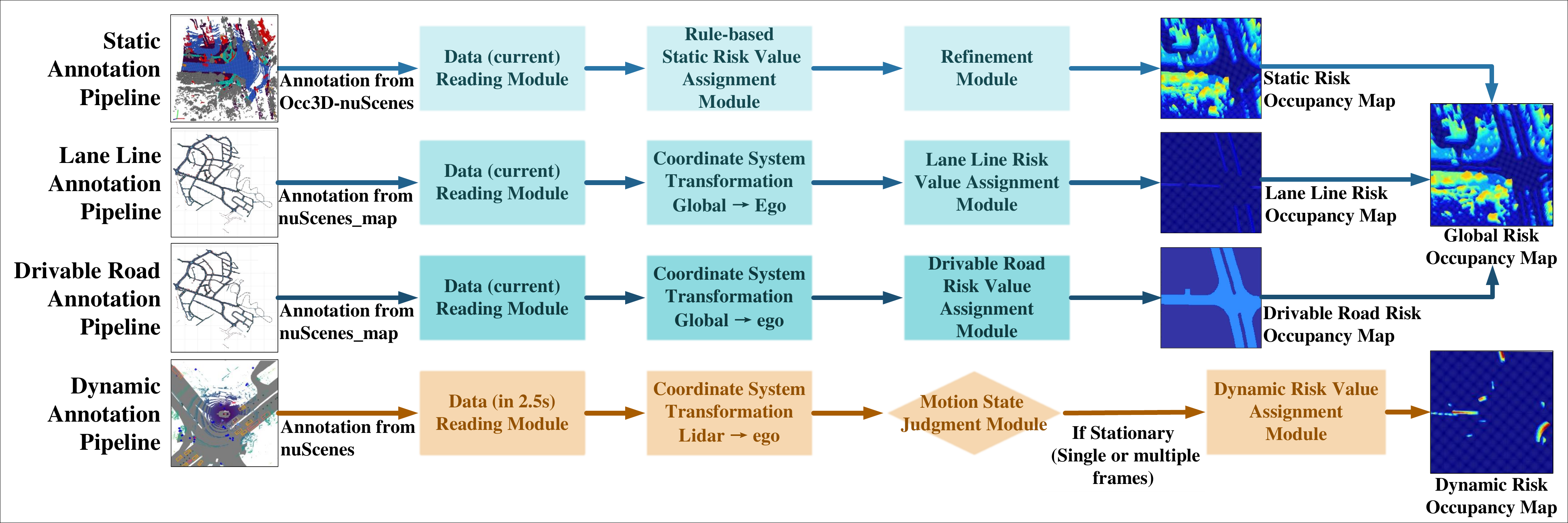}
    \caption{The four automated annotation pipelines building RiskOcc4D-nuScenes, fusing static, regulatory, drivable-area, and dynamic-future cues into the global and dynamic risk-aware occupancy maps.}
    \label{fig:pipelines}
\end{figure*}

\noindent\textbf{Temporal Query Consistency.} TQC calibrates query reliability before motion self-attention without learnable parameters. It caches previous-frame agent states, aligns them to the current ego frame via CAN-bus ego-motion, and matches each query to its nearest predecessor. A consistency weight combines feature cosine similarity with a distance-decayed spatial score, using a fallback for newly appearing objects and a lower bound to avoid hard deletion. The weights form a pairwise geometric-mean gate that is mapped to a log-odds bias $\beta$ and applied through a stop-gradient $\mathrm{sg}(\cdot)$: $\mathrm{Attn}(Q,K,V)=\mathrm{softmax}(QK^{\top}/\sqrt{d}+\mathrm{sg}(\beta))V$. This mechanism amplifies consistent tokens and suppresses spurious or flickering ones. Because TQC is inference-only and gradient-free, it remains lightweight and falls back to the base model on the first frame.

\noindent\textbf{Generative planning.} The instance tokens, densified by RiskOcc, reweighted by TQC, and initialized at the ego slot from historical ego states \cite{li2023fb}, are encoded by a VAE into a latent Gaussian $p(z\mid T)\sim\mathcal{N}(\mu_i,\sigma_i)$ \cite{genad}; a sampled $z$ is decoded by a GRU and an MLP into multimodal ego and agent trajectories.

\section{RiskOcc4D-nuScenes Dataset}
\label{sec:dataset}

Conventional 3D occupancy datasets contain substantial spatial redundancy, use semantics that are not directly aligned with E2E planning, and lack unified collision-regulatory risk encoding and temporal risk annotations. We address these limitations with \textit{RiskOcc4D-nuScenes}, built upon nuScenes and Occ3D-nuScenes. To our knowledge, it is the first 4D risk-aware occupancy benchmark designed for E2E autonomous driving. Four automated pipelines organize static obstacle risk, traffic-regulation risk, and dynamic future risk within a common annotation framework that remains compatible with representative 3D occupancy architectures.

\subsection{Task Definition}

Combining the avoidance-priority principle with Bayesian risk decision theory, we instantiate the abstract optimal representation $R^*$ in Eq.~(\ref{eq1}) by minimizing the bottleneck risk that aligns perception with planning:
\begin{equation}
  \mathcal{L}(T_{\mathrm{plan}} \mid Z)
  = \mathbb{E}\!\left[\max\!\left(
      \ell_{\mathrm{col}}(T_{\mathrm{plan}}, O),\;
      \ell_{\mathrm{reg}}(T_{\mathrm{plan}}, M)
    \right) \mid Z\right],
\end{equation}
where $Z$ is the multi-view input, $(\ell_{\mathrm{col}}, O)$ the collision risk and obstacle set, and $(\ell_{\mathrm{reg}}, M)$ the regulatory risk and traffic-rule constraints. The $\max$ operator enforces hard-constraint alignment, avoiding accumulation overflow and inter-term gradient conflict, thereby providing the supervisory signal for the risk-aware occupancy maps in Section~\ref{data pipe}.

\noindent\textbf{Input.}\quad
Six surround-view camera sequences under nuScenes calibration: intrinsics $\{K_i\}$, extrinsics $\{[R_i \mid t_i]\}$, and $N$ consecutive frames $\{I_{i,t} \in \mathbb{R}^{H_i \times W_i \times 3}\}$ ($i{=}1,\dots,6$; $t{=}1,\dots,N$).

\begin{table*}[t!]
\centering

\fontsize{9}{11}\selectfont
\renewcommand{\arraystretch}{1.3}
\setlength{\tabcolsep}{2pt}
\begin{tabular}{lcccccclclc}
\hline
\multirow{3}{*}{\textbf{Baselines}} & \multirow{3}{*}{\textbf{Release}} & \multicolumn{9}{c}{\textbf{Metrics}} \\ \cline{3-11}
 & & \multicolumn{5}{c}{\textbf{Global}} & & \textbf{Dynamic} & & \textbf{Infer.\ Speed} \\ \cline{3-7} \cline{9-9} \cline{11-11}
 & & \textbf{mIoU} $\uparrow$ & \textbf{MAE} $\downarrow$ & \textbf{MAE@15} $\downarrow$ & \textbf{MAE@30} $\downarrow$ & \textbf{SSIM} $\uparrow$ & & \textbf{Mask-MAE} $\downarrow$ & & \textbf{FPS} $\uparrow$ \\ \hline
\rowcolor[HTML]{F0F0F0}PanoOcc-S \cite{wang2024panoocc} & CVPR 24 & 14.54 & 0.121 & 0.022 & 0.079 & 0.131 & & 0.732 & & 6.0 \\
PanoOcc \cite{wang2024panoocc} & CVPR 24 & 14.71 & 0.121 & 0.022 & 0.079 & 0.131 & & 0.731 & & 1.9 \\
\rowcolor[HTML]{F0F0F0}OccProphet \cite{chen2025occprophet} & ICLR 25 & 17.39 & 0.099 & 0.017 & 0.067 & 0.355 & & \textbf{0.524} & & 0.7 \\
ViewFormerOcc \cite{viewformer} & ECCV 24 & 21.38 & 0.093 & 0.015 & 0.060 & 0.379 & & 0.734 & & 19.0 \\
\rowcolor[HTML]{F0F0F0}FO(BEVDET) \cite{yu2023flashocc} & ICCV 23 & 22.03 & 0.088 & 0.014 & 0.058 & 0.424 & & 0.682 & & \textbf{66.2} \\
BEVFormer-Occ \cite{li2022bevformer} & ECCV 22 & 23.76 & 0.086 & 0.014 & 0.057 & 0.437 & & 0.565 & & 11.1 \\
\rowcolor[HTML]{F0F0F0}CVTOcc \cite{CVT} & ECCV 24 & 24.43 & 0.094 & 0.016 & 0.063 & 0.385 & & 0.623 & & 2.1 \\
FO(Stereo4D) \cite{yu2023flashocc} & ICCV 23 & 25.14 & 0.082 & 0.011 & 0.052 & 0.443 & & 0.664 & & 10.4 \\
\rowcolor[HTML]{F0F0F0}DHDOcc-S \cite{dhd} & ICRA 25 & 25.45 & 0.092 & 0.015 & 0.061 & 0.399 & & 0.610 & & 18.7 \\
ProtoOcc \cite{kim2025protoocc} & AAAI 25 & 28.58 & 0.082 & 0.012 & 0.053 & 0.422 & & 0.589 & & 7.1 \\
\rowcolor[HTML]{F0F0F0}FB-Occ \cite{li2023fb} & ICCV 23 & 32.96 & \textbf{0.075} & \textbf{0.010} & \textbf{0.048} & \textbf{0.480} & & 0.675 & & 34.1 \\
DHDOcc-M \cite{dhd} & ICRA 25 & \textbf{34.15} & 0.079 & 0.011 & 0.050 & 0.455 & & 0.559 & & 3.5 \\ \hline
\end{tabular}
\caption{Feasibility benchmark for predicting risk-aware occupancy with representative 3D occupancy backbones. Best values are shown in \textbf{bold}.}
\label{tab:comparison}
\end{table*}

\noindent\textbf{Output.}\quad
Ego-centric BEV risk-aware occupancy maps: a global map $RM_t \in \mathbb{R}^{H \times W}$, where occupancy encodes spatial location and risk values quantify hazard attention carrying collision and regulatory information, and a dynamic map $RM_d$ that fuses future steps $\Delta t$ to encode motion-prediction uncertainty. They provide three targets: (1)~discrete global risk classification, (2)~continuous global risk regression, and (3)~dynamic temporal risk prediction.

\subsection{Dataset Construction Pipelines}
\label{data pipe}
Our four-stage automated framework generates the global and dynamic risk-aware occupancy maps (Figure~\ref{fig:pipelines}), compressing redundant 3D representations while enriching label completeness.

\subsubsection{Static Annotation Pipeline (Pipeline-1)}
This pipeline extracts static risk priors from Occ3D-nuScenes via 3D-to-BEV reduction. For each $x$-$y$ cell, vertical marginalization within the critical height range $H_{\mathrm{crit}}\in[0,\,2.2]\,\mathrm{m}$ projects semantic and geometric cues onto the BEV plane, where the driving risk attention $V_{\mathrm{risk}}$ is a joint function of local height distribution and semantic category priors. To suppress quantization artifacts and smooth jagged boundaries, we refine the initial risk-aware occupancy $\mathbf{R}_{\mathrm{init}}$ via morphological dilation-erosion and a custom convolution-kernel infilling:
$
\hat{\mathbf{R}}_0 = \mathcal{F}\!\left((\mathbf{R}_{\mathrm{init}}\oplus\mathbf{K}_d)\ominus\mathbf{K}_e\right),
$
where $\oplus$ and $\ominus$ denote dilation and erosion with structuring elements $\mathbf{K}_d$ and $\mathbf{K}_e$, and $\mathcal{F}(\cdot)$ propagates neighboring context to repair sparse regions.

\subsubsection{Lane Line Annotation Pipeline (Pipeline-2)}
This pipeline extracts an ego-centric $80\,\mathrm{m}{\times}80\,\mathrm{m}$ semantic map from nuScenes and injects regulatory priors by assigning $V_{\mathrm{risk}}=0.1$ to road-marking cells (crosswalks, stop lines, lane dividers). This explicit rule encoding prunes the trajectory hypothesis space, reduces decision entropy, and sets a low-risk regulatory baseline in the hierarchical risk framework.

\subsubsection{Drivable Road Annotation Pipeline (Pipeline-3)}
To counter missing drivable-area labels and LiDAR-sparsity noise in Occ3D-nuScenes, high-fidelity drivable masks from nuScenes HD maps are fused with the Pipeline-1/2 outputs through deterministic spatial fusion, completing undefined areas and removing outliers (Figure~\ref{fig:pipelines}). For occluded blind-spot cells with unobservable occupancy, a small positive uncertainty prior of $0.05$---below regulatory and collision risk---encourages cautious attention until later viewpoints provide definitive evidence. Fusing Pipelines~1--3 yields the global risk-aware occupancy map, stored as both classification and regression labels.

\subsubsection{Dynamic Annotation Pipeline (Pipeline-4)}
This pipeline builds the dynamic risk potential field from observable temporal occupancy rather than unreliable kinematic fitting. Moving objects ($v>0.5\,\mathrm{m/s}$) are tracked across frames to record actual occupied positions over the next $2.5\,\mathrm{s}$ at $0.5\,\mathrm{s}$ intervals. Risk attention is inversely proportional to the time step---near-future moments carry higher attention, following the ``urgent-near, relaxed-far'' defensive-driving logic---yielding an initial field $\hat{\mathbf{R}}_0^{\mathrm{dyn}}$. A spatial Gaussian diffusion $\mathbf{R}^{\mathrm{dyn}} = \mathcal{G}_\sigma * \hat{\mathbf{R}}_0^{\mathrm{dyn}}$ (kernel $\mathcal{G}_\sigma$, $*$ spatial convolution) produces a continuous field with Lipschitz-continuous variation, modeling risk attenuation toward free space. Per Section~\ref{sec:paradigm}, dynamic objects take $V_{\mathrm{risk}}^{\mathrm{dyn}} \in [0.5,\,1.0]$ and are stored separately from static labels to avoid multi-task interference.

\section{Experiments}
\label{sec:experiments}

Multiple 3D occupancy prediction methods are evaluated on RiskOcc4D-nuScenes. Experiments are further conducted to validate the effectiveness and marginal gains of this representation, and the downstream utility of predicted risk maps for E2E driving is assessed.

\subsection{Experimental Setup}
\label{subsec:setup}

\subsubsection{Implementation details}
All occupancy baselines use their default configurations with only the output heads replaced: the classification head retains each method's original supervision, while both regression heads use the Huber loss~\cite{huber}. Training runs on multiple NVIDIA A100 GPUs; inference and benchmarking use an Intel i9-13900K CPU with an NVIDIA RTX 4090 GPU.

\subsubsection{Evaluation Metrics}
\label{sec:5.1}
We adopt a four-fold protocol. \textit{(1)~Classification} uses mIoU (mean Jaccard index)~\cite{everingham2010pascal} for risk-category alignment; \textit{(2)~Regression} uses MAE ($L_1$), with MAE@15/MAE@30 restricted to ego-centric $15{\times}15$ and $30{\times}30\,\text{m}^2$ regions to emphasize the safety-critical near field; \textit{(3)~Structural fidelity} uses SSIM~\cite{wang2004image} for luminance, contrast, and structural consistency beyond pointwise accuracy. \textit{(4)~Dynamic prediction} uses Mask-MAE, which counters the extreme imbalance from dominant background zeros via a binary mask $M_i = \mathbf{1}[P_i > \tau \;\text{or}\; G_i > \tau]$ ($\tau = 0.1$):
\begin{equation}
    \text{Mask-MAE} = \tfrac{1}{\sum_i M_i} \sum_{i=1}^{N} M_i |P_i - G_i|,
\end{equation}
restricting evaluation to the support $\max(P_i, G_i) > \tau$ and improving discriminability.

For end-to-end planning, we distinguish the two commonly used protocols explicitly. ST-P3~\cite{stp3} reports temporally averaged displacement error (TemAvg), averaging the instantaneous errors at all 0.5\,s steps up to each horizon, whereas UniAD~\cite{uniad} reports the endpoint displacement error (NoAvg) at the queried horizon. We assign all internally evaluated planning results to these protocols according to these definitions.

\begin{figure*}[h!]
    \centering
    \includegraphics[width=0.99\linewidth, trim=2 2 2 2, clip]{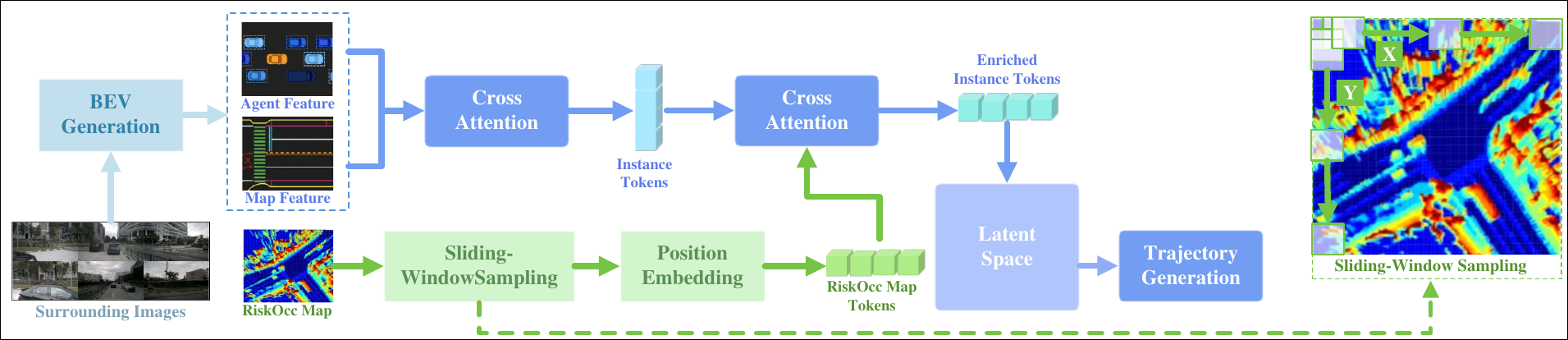}
    \caption{Architecture of the RiskOcc-GenAD fusion experiment: the predicted risk-aware occupancy map is sampled by a sliding window and fused with GenAD's instance tokens via cross-attention, entering as an additional input while GenAD's protocols stay unchanged.}
    \label{fig:closeloop}
\end{figure*}

\subsection{Benchmarking on RiskOcc4D-nuScenes}
\label{subsec:baseline}

We migrate a broad set of 3D occupancy methods to RiskOcc4D-nuScenes---PanoOcc~\cite{wang2024panoocc}, OccProphet~\cite{chen2025occprophet}, BEVDet4DOcc~\cite{huang2022bevdet4d}, ViewFormerOcc~\cite{viewformer}, FlashOcc(FO)~\cite{yu2023flashocc}, BEVFormer-Occ~\cite{li2022bevformer}, CVTOcc*~\cite{CVT}, ProtoOcc~\cite{kim2025protoocc}, FB-Occ~\cite{li2023fb}, and DHDOcc~\cite{dhd}---with results in Table~\ref{tab:comparison}.

\begin{table}[t!]

\centering
\fontsize{9}{11}\selectfont
\setlength{\tabcolsep}{3.2pt}
\renewcommand{\arraystretch}{1.1}
\begin{tabular}{lcccclcccc}
\hline
\multirow{2}{*}{\textbf{Methods}} & \multicolumn{4}{c}{\textbf{ST-P3 Metrics} $\downarrow$} & & \multicolumn{4}{c}{\textbf{UniAD Metrics} $\downarrow$} \\ \cline{2-5} \cline{7-10}
 & \textbf{1~s} & \textbf{2~s} & \textbf{3~s} & \cellcolor[HTML]{DCDCDC}\textbf{Avg.} & & \textbf{1~s} & \textbf{2~s} & \textbf{3~s} & \cellcolor[HTML]{DCDCDC}\textbf{Avg.} \\ \hline
ST-P3              & 0.23          & 0.62          & 1.27          & \cellcolor[HTML]{DCDCDC}0.71          & & -             & -             & -             & \cellcolor[HTML]{DCDCDC}-             \\
OccNet               & 0.21          & 0.59          & 1.37          & \cellcolor[HTML]{DCDCDC}0.72          & & -             & -             & -             & \cellcolor[HTML]{DCDCDC}-             \\
VAD-Tiny       & 0.21          & 0.35          & 0.58          & \cellcolor[HTML]{DCDCDC}0.38          & & -             & -             & -             & \cellcolor[HTML]{DCDCDC}-             \\
VAD-Base      & 0.07          & 0.17          & 0.41          & \cellcolor[HTML]{DCDCDC}0.22          & & 0.01          & 0.30          & 0.95          & \cellcolor[HTML]{DCDCDC}0.42          \\
FusionAD    & 0.25          & 0.13          & 0.25 & \cellcolor[HTML]{DCDCDC}0.21          & & -             & -             & -             & \cellcolor[HTML]{DCDCDC}-             \\
UniAD         & -             & -             & -             & \cellcolor[HTML]{DCDCDC}-             & & 0.05          & 0.17          & 0.71          & \cellcolor[HTML]{DCDCDC}0.31          \\
PPAD (P.)      & 0.08          & 0.12          & 0.38          & \cellcolor[HTML]{DCDCDC}0.19          & & 0.02          & 0.20          & 0.93          & \cellcolor[HTML]{DCDCDC}0.38          \\
PPAD          & 0.07          & 0.15          & 0.36          & \cellcolor[HTML]{DCDCDC}0.19          & & 0.03          & 0.22          & 0.73          & \cellcolor[HTML]{DCDCDC}0.33          \\
PARADrive     & -             & -             & -             & \cellcolor[HTML]{DCDCDC}-             & & 0.07          & 0.25          & \textbf{0.60} & \cellcolor[HTML]{DCDCDC}0.30          \\
AD-MLP   & 0.17          & 0.18          & 0.24          & \cellcolor[HTML]{DCDCDC}0.20          & & -             & -             & -             & \cellcolor[HTML]{DCDCDC}-             \\
\rowcolor[HTML]{F0F0F0} \textbf{GenAD}  & 0.06          & 0.15          & 0.32          & \cellcolor[HTML]{DCDCDC}0.17          & & 0.05          & 0.29          & 0.87          & \cellcolor[HTML]{DCDCDC}0.40          \\
\rowcolor[HTML]{F0F0F0} \textbf{Ours}                & \textbf{0.02} & \textbf{0.04} & \textbf{0.19} & \cellcolor[HTML]{DCDCDC}\textbf{0.08} & & \textbf{0.00} & \textbf{0.14} & 0.64 & \cellcolor[HTML]{DCDCDC}\textbf{0.26} \\ \hline
\end{tabular}
\caption{Collision rate (\%) comparison among E2E methods. Ours = GenAD~\cite{genad}\,$+$\,RiskOcc Map; P.\,=\,progressive training. Best values are shown in \textbf{bold}.}
\label{tab:collision}
\end{table}

\textbf{Global risk-aware occupancy.}\quad
DHDOcc-M~\cite{dhd} leads in mIoU (34.15), indicating strong risk-category discrimination, while FB-Occ~\cite{li2023fb} attains the best regression at all scales (MAE 0.075, MAE@15 0.010, MAE@30 0.048) and the highest SSIM (0.480), best preserving the ground-truth spatial gradients.

\textbf{Dynamic risk-aware occupancy.}\quad
OccProphet achieves the best Mask-MAE (0.524), followed by DHDOcc-M (0.559) and BEVFormer-Occ (0.565). BEVDet4DOcc is an outlier (1.856), likely because of distribution shift in its dynamic head; the remaining baselines fall within 0.559--0.734. The generally limited performance highlights the challenge of multi-step risk propagation.

\textbf{Inference efficiency.}\quad
FO(BEVDET)~\cite{yu2023flashocc} reaches 66.2\,FPS without temporal history but with limited accuracy, whereas FB-Occ~\cite{li2023fb} offers a strong speed-accuracy trade-off (34.1\,FPS) alongside the strongest global regression metrics.

\begin{table}[t!]
\centering

\fontsize{9}{11}\selectfont
\setlength{\tabcolsep}{2.8pt}
\renewcommand{\arraystretch}{1.35}
\begin{tabular}{lcccclcccc}
\hline
\multirow{2}{*}{\textbf{Inputs}} & \multicolumn{4}{c}{\textbf{ST-P3 Metrics} $\downarrow$} & & \multicolumn{4}{c}{\textbf{UniAD Metrics} $\downarrow$} \\ \cline{2-5} \cline{7-10}
 & \textbf{1~s} & \textbf{2~s} & \textbf{3~s} & \cellcolor[HTML]{DCDCDC}\textbf{Avg.} & & \textbf{1~s} & \textbf{2~s} & \textbf{3~s} & \cellcolor[HTML]{DCDCDC}\textbf{Avg.} \\ \hline
\rowcolor[HTML]{F0F0F0} \textbf{RiskOcc Map} & \textbf{0.02} & \textbf{0.04} & \textbf{0.19} & \cellcolor[HTML]{DCDCDC}\textbf{0.08} & & \textbf{0.00} & \textbf{0.14} & \textbf{0.64} & \cellcolor[HTML]{DCDCDC}\textbf{0.26} \\
3D-Occ         & 0.09          & 0.18          & 0.33          & \cellcolor[HTML]{DCDCDC}0.20          & & 0.08          & 0.29          & 0.74          & \cellcolor[HTML]{DCDCDC}0.37          \\
Binary Map     & 0.11          & 0.21          & 0.39          & \cellcolor[HTML]{DCDCDC}0.24          & & 0.14          & 0.39          & 0.90          & \cellcolor[HTML]{DCDCDC}0.48          \\
Gaussian Map   & 0.21          & 0.32          & 0.37          & \cellcolor[HTML]{DCDCDC}0.33          & & 0.27          & 0.51          & 0.86          & \cellcolor[HTML]{DCDCDC}0.55          \\ \hline
\end{tabular}
\caption{Ablation of intermediate representations on collision rate (\%). Best in \textbf{bold}.}
\label{tab:ablation}
\end{table}


\subsection{End-to-End Autonomous Driving Integration}
\label{subsec:e2e}

\subsubsection{Experimental Design}
We integrate the risk-aware occupancy map into GenAD~\cite{genad}, an E2E planner encoding dynamic agents and road topology as instance and map tokens. Such discrete object-level representations are spatially incomplete, capturing only detected targets while lacking explicit encoding of dense spatial risk, long-tail obstacles, and ambiguous occupied regions---gaps that our dense, continuous risk distribution complements. Let $\mathcal{Z}_{\text{inst+map}}$ and $\mathcal{Z}_{\text{risk}}$ denote the joint instance-and-map and the risk representations. Our central hypothesis is a positive conditional mutual information gain:
\begin{equation}
\label{eq:mi}
  \Delta I = I\!\left(\mathcal{A}_{\text{safe}};\,
    \mathcal{Z}_{\text{risk}} \mid
    \mathcal{Z}_{\text{inst+map}}\right) > 0,
\end{equation}
i.e., the risk representation carries safety-relevant information not captured by instance and map tokens. The risk map is sampled via a sliding window and fused with GenAD's instance tokens through cross-attention (Figure~\ref{fig:closeloop}), allowing each instance token to attend to surrounding grid-level risk. GenAD's training and evaluation settings remain unchanged, while the risk map enters solely as an additional input, isolating the contribution of the proposed representation.

\begin{table*}[t!]
\centering

\fontsize{9}{11}\selectfont
\renewcommand{\arraystretch}{1.1}
\setlength{\tabcolsep}{4.4pt}
\begin{tabular}{@{}ccccccccccccccccccccccc@{}}
\toprule
\multicolumn{3}{c}{\textbf{Methods}}                                                                                        &  & \multicolumn{9}{c}{\textbf{UniAD Metrics}}                                                                                                                                                                                                                                                                               &  & \multicolumn{9}{c}{\textbf{ST-P3 Metrics}}                                                                                                                                                                                                                                                                               \\ \cmidrule(r){1-3} \cmidrule(lr){5-13} \cmidrule(l){15-23} 
\multicolumn{1}{c}{\multirow{2}{*}{\textbf{M1}}} & \multicolumn{1}{c}{\multirow{2}{*}{\textbf{M2}}} & \multicolumn{1}{c}{\multirow{2}{*}{\textbf{M3}}} &  & \multicolumn{4}{c}{\textbf{L2 ($\text{m}$) $\downarrow$}}                                                                                                 &  & \multicolumn{4}{c}{\textbf{Collision Rate (\%) $\downarrow$}}                                                                                             &  & \multicolumn{4}{c}{\textbf{L2 ($\text{m}$) $\downarrow$}}                                                                                                 &  & \multicolumn{4}{c}{\textbf{Collision Rate (\%) $\downarrow$}}                                                                                             \\ \cmidrule(lr){5-8} \cmidrule(lr){10-13} \cmidrule(lr){15-18} \cmidrule(l){20-23} 
\multicolumn{1}{c}{}                    & \multicolumn{1}{c}{}                    & \multicolumn{1}{c}{}                    &  & \multicolumn{1}{c}{\textbf{1 s}} & \multicolumn{1}{c}{\textbf{2 s}} & \multicolumn{1}{c}{\textbf{3 s}} & \multicolumn{1}{c}{\cellcolor[HTML]{DCDCDC}Avg.}       &  & \multicolumn{1}{c}{\textbf{1 s}} & \multicolumn{1}{c}{\textbf{2 s}} & \multicolumn{1}{c}{\textbf{3 s}} & \multicolumn{1}{c}{\cellcolor[HTML]{DCDCDC}Avg.}       &  & \multicolumn{1}{c}{\textbf{1 s}} & \multicolumn{1}{c}{\textbf{2 s}} & \multicolumn{1}{c}{\textbf{3 s}} & \multicolumn{1}{c}{\cellcolor[HTML]{DCDCDC}Avg.}       &  & \multicolumn{1}{c}{\textbf{1 s}} & \multicolumn{1}{c}{\textbf{2 s}} & \multicolumn{1}{c}{\textbf{3 s}} & \multicolumn{1}{c}{\cellcolor[HTML]{DCDCDC}Avg.}       \\ \midrule
\multicolumn{1}{c}{\checkmark}               &                                         &                                         &  & 0.38                    & 0.85                    & 1.52                    & \cellcolor[HTML]{DCDCDC}0.92                           &  & 0.07           & \textbf{0.07}           & \textbf{0.64}           & \cellcolor[HTML]{DCDCDC}\textbf{0.26} &  & 0.29                    & 0.51                    & 0.78                    & \cellcolor[HTML]{DCDCDC}0.52                           &  & 0.08           & 0.07           & \textbf{0.21}           & \multicolumn{1}{c}{\cellcolor[HTML]{DCDCDC}0.12} \\
\multicolumn{1}{c}{\checkmark}               & \multicolumn{1}{c}{\checkmark}               &                                         &  & \textbf{0.21}                    & \textbf{0.58}                    & \textbf{1.22}           & \cellcolor[HTML]{DCDCDC}\textbf{0.67}                           &  & 0.08                    & 0.16                    & 0.84                    & \cellcolor[HTML]{DCDCDC}0.36                           &  & \textbf{0.16}                    & \textbf{0.31}           & \textbf{0.56}           & \cellcolor[HTML]{DCDCDC}\textbf{0.34} &  & 0.08           & 0.10                    & 0.27                    & \multicolumn{1}{c}{\cellcolor[HTML]{DCDCDC}0.15}                           \\
\multicolumn{1}{c}{\checkmark}               & \multicolumn{1}{c}{\checkmark}               & \multicolumn{1}{c}{\checkmark}               &  & \textbf{0.21}           & 0.59           & 1.26                    & \cellcolor[HTML]{DCDCDC}0.69 &  & \textbf{0.00}           & 0.14                    & 0.86                    & \cellcolor[HTML]{DCDCDC}0.33 &  & \textbf{0.16}           & 0.32           & 0.57           & \cellcolor[HTML]{DCDCDC}0.35 &  & \textbf{0.00}                    & \textbf{0.04}                    & \textbf{0.21}                    & \multicolumn{1}{c}{\cellcolor[HTML]{DCDCDC}\textbf{0.08}}                           \\ 
\bottomrule
\end{tabular}
\caption{ROIDrive framework ablation. M1: risk-aware occupancy perception; M2: ego-token prior initialization; M3: Temporal Query Consistency (TQC). Best values are shown in \textbf{bold}.}
\label{tab:ablation1}
\end{table*}

\subsubsection{Comparative Results}
Table~\ref{tab:collision} reports collision rates on the nuScenes validation set. Relative to GenAD, our method reduces the average collision rate by 35.0\% under UniAD (0.40\%\,$\to$\,0.26\%) and by 52.9\% under ST-P3 (0.17\%\,$\to$\,0.08\%). The largest relative gains occur at the 2\,s horizon (51.7\% under UniAD and 73.3\% under ST-P3), where accumulated trajectory uncertainty makes dense spatial risk priors particularly valuable. Our method records zero collisions at 1\,s under UniAD and the lowest 1\,s collision rate under ST-P3 (0.02\%), while achieving the lowest average collision rate under both metrics among the compared methods.

\noindent\textbf{Overall performance.} On the nuScenes \cite{caesar2020nuscenes} validation set, ROIDrive attains an average L2 error and collision rate of $0.69$\,m / $0.33\%$ under the UniAD metric and $0.35$\,m / $0.08\%$ under the ST-P3 metric, with the lowest average collision rate under both protocols among the compared methods.

\begin{table}[t!]

\centering
\fontsize{9}{11}\selectfont
\renewcommand{\arraystretch}{1.1}
\setlength{\tabcolsep}{3.1pt}
\begin{tabular}{@{}cc@{\hspace{0.1pt}}cccc@{\hspace{0.1pt}}cccc@{}}
\toprule
\multicolumn{2}{c}{\textbf{Components}}                                                                                         & \multicolumn{4}{c}{\textbf{UniAD Collision Rate $\downarrow$}}  & \multicolumn{4}{c}{\textbf{ST-P3 Collision Rate $\downarrow$}}                                                                                                                                                                                                                                                                        \\ \cmidrule(r){1-2} \cmidrule(lr){3-6} \cmidrule(lr){7-10} 
\multicolumn{1}{c}{\textbf{C1}} & \multicolumn{1}{c}{\textbf{C2}}    & \multicolumn{1}{c}{\textbf{1 s}} & \multicolumn{1}{c}{\textbf{2 s}} & \multicolumn{1}{c}{\textbf{3 s}} & \multicolumn{1}{c}{%
\cellcolor[HTML]{DCDCDC}
\makebox[4pt][c]{Avg.}}         & \multicolumn{1}{c}{\textbf{1 s}} & \multicolumn{1}{c}{\textbf{2 s}} & \multicolumn{1}{c}{\textbf{3 s}} & \multicolumn{1}{c}{\cellcolor[HTML]{DCDCDC}Avg.}               \\ \midrule
\multicolumn{1}{c}{\checkmark}               &                                                                     & 0.08                    & 0.18                    & 0.84                    & \multicolumn{1}{c}{\cellcolor[HTML]{DCDCDC}0.37}                             & 0.08           & 0.11           & 0.30           & \multicolumn{1}{c}{\cellcolor[HTML]{DCDCDC}0.16}  \\
                & \multicolumn{1}{c}{\checkmark}                                              & \textbf{0.06}                    & 0.27           & \textbf{0.54}           & \multicolumn{1}{c}{\cellcolor[HTML]{DCDCDC}0.29}   & \textbf{0.05}           & 0.12                    & 0.22                   & \multicolumn{1}{c}{\cellcolor[HTML]{DCDCDC}0.13}                                                      \\
\multicolumn{1}{c}{\checkmark}               & \multicolumn{1}{c}{\checkmark}                             & 0.07           & \textbf{0.07}           & 0.64           & \multicolumn{1}{c}{\cellcolor[HTML]{DCDCDC}\textbf{0.26}}  & 0.08                    & \textbf{0.07}                    & \textbf{0.21}                    & \multicolumn{1}{c}{\cellcolor[HTML]{DCDCDC}\textbf{0.12}}                                            \\ 
\bottomrule
\end{tabular}
\caption{RiskOcc-branch component ablation without ego-token initialization or TQC. C1: Bi-Mamba2; C2: SDMap+ fusion. Best values are shown in \textbf{bold}.}
\label{tab:ablation2}
\end{table}

\subsubsection{Ablation Study}
We evaluate the contributions of the framework modules, the occupancy-branch components, and alternative intermediate representations while keeping all unrelated settings fixed.

\noindent\textbf{Ablation: framework modules.} Table~\ref{tab:ablation1} incrementally adds the modules. RiskOcc alone (M1) sets a dense risk prior and a safe planning baseline; ego-token initialization (M2) adds the ego motion prior and sharply lowers average L2 (UniAD $0.92\!\to\!0.67$, ST-P3 $0.52\!\to\!0.34$) at marginal collision cost; TQC (M3) then reduces collisions relative to the ego-initialized variant (UniAD $0.36\%\!\to\!0.33\%$, ST-P3 $0.15\%\!\to\!0.08\%$) by down-weighting unreliable queries while keeping L2 nearly unchanged, achieving the intended safety-accuracy balance.

\noindent\textbf{Ablation: occupancy components.} Built on the RiskOcc-only model without ego-token initialization or TQC, Table~\ref{tab:ablation2} isolates the two components of the branch. The Bi-Mamba2 backbone (C1) provides dynamic avoidance from onboard vision, whereas the SDMap+ prior (C2) fills static blind spots and most reduces the $3$\,s collision rate; enabling both yields the lowest average collision rate (UniAD $0.26\%$, ST-P3 $0.12\%$), confirming that dynamic perception and static priors are complementary.

\textbf{3D-Occ.}\quad
Conventional 3D semantic voxels are adaptively pooled to the RiskOcc Map size and fused identically. Gains are mixed---slightly better than baseline under UniAD (0.37 vs.\ 0.40) but worse under ST-P3 (0.20 vs.\ 0.17): although raw voxels carry high mutual information, adaptive pooling lowers the effective information rate and planning-irrelevant semantic redundancy adds noise. By distilling safety-critical information at construction time, RiskOcc attains higher task-relevant density in a lower-dimensional BEV, with consistent gains under both protocols (UniAD/ST-P3: 0.26\,/\,0.08).

\textbf{Binary Map.}\quad
Collapsing continuous risk to a single bit (risk\,=\,1, no risk\,=\,0) erases ordinal relationships among threat levels. The resulting rates (UniAD Avg. 0.48; ST-P3 Avg. 0.24) are substantially higher than those of RiskOcc (0.26\,/\,0.08) and also exceed the no-map baseline (0.40\,/\,0.17), where each pair is ordered as UniAD/ST-P3. This result indicates that a binary signal without risk gradients can mislead the planner. Hence risk stratification, rather than mere spatial occupancy, is central to the observed safety gains.

\textbf{Gaussian Map.}\quad
As a control, a random Gaussian map of identical size but carrying no risk semantics yields the highest collision rates across all metrics (UniAD Avg. 0.55; ST-P3 Avg. 0.33), performing worse than the no-map baseline. This result rules out improvements from the additional cross-attention parameters alone and supports the importance of semantically meaningful risk encoding.

Together, these ablations rule out network capacity, high-dimensional semantics, and binary occupancy as alternative explanations, providing evidence consistent with---though not a direct measurement of---the conditional mutual information gain $\Delta I > 0$ in Eq.~(\ref{eq:mi}).

\FloatBarrier
\section{Conclusion}
\label{sec:conclusion}

This paper introduced risk-aware occupancy as a dense perception
representation for safety-oriented E2E autonomous driving. We constructed
RiskOcc4D-nuScenes on top of Occ3D-nuScenes and benchmarked representative
3D occupancy architectures to evaluate the learnability and transferability
of the proposed representation. Integrating predicted risk-aware occupancy
into GenAD~\cite{genad} through sliding-window sampling reduces collision
rates by 35.0\% under UniAD and 52.9\% under ST-P3. We further developed
ROIDrive, an E2E framework that combines an independent risk-aware occupancy
branch with temporal query consistency for safety-oriented planning.

    \begin{small}
    \bibliography{aaai2027}
    \end{small}

\end{document}


\maketitle

\section{Appendix Overview}

This appendix complements the main paper with mathematical details that are omitted from the main text due to space constraints. We focus on four aspects: the formal construction of risk-aware occupancy, the annotation equations behind RiskOcc4D-nuScenes, the ROIDrive fusion modules, and the full open-loop planning table with both L2 and collision metrics. Qualitative figures and ablation tables that duplicate the main paper are omitted.

\section{A. Risk-Aware Occupancy Formulation}

\subsection{Rationale of Risk Value Assignment}

Risk-aware occupancy is not intended to provide a physically calibrated accident probability or a unique absolute risk value. Instead, it serves as an ordinal, planner-facing attention prior: cells that are more likely to constrain safe motion should receive larger values than cells that are less relevant to planning. We therefore use a simple monotone assignment rather than claiming that the exact numeric values are optimal. The assignment is based on reliably observable cues, including collision-relevant height, semantic category, HD-map elements, visibility, and future agent footprints. Latent attributes such as object mass are not used, because they are not reliably recoverable from perception or annotation. Under this view, the important invariant is the relative ordering among risk sources, from uncertainty and regulatory markings to static obstacles and near-future dynamic agents; the numeric intervals act as normalized supervision for the network and keep label construction consistent with sensor-based prediction.

The main paper defines risk-aware occupancy as a planning-oriented dense representation. Let $\mathcal{V}$ denote the original 3D occupancy volume and let $T_{\mathrm{plan}}$ denote the target planning trajectory. The representation objective can be written as:
\begin{equation}
  \begin{aligned}
  R^{*} &= \arg\max_R I(R;T_{\mathrm{plan}}),\\
  \mathrm{s.t.}\quad
  \dim(R)&\ll \dim(\mathcal{V}).
  \end{aligned}
  \label{eq:supp_minimal_stat}
\end{equation}
where $I(\cdot;\cdot)$ denotes mutual information. This formulation states that the BEV risk field should retain information relevant to planning while discarding height and semantic redundancy that is not needed by the planner.

For each BEV cell $(x,y)$, the relevant vertical interval is restricted to the collision-critical range
$H_{\mathrm{crit}}=[0,2.2]\,\mathrm{m}$. Let $\mathcal{Z}_{x,y}$ be the set of occupied heights in this interval:
\begin{equation}
  \begin{aligned}
  \mathcal{Z}_{x,y}
  = \{z \mid
  &\mathcal{V}(x,y,z)=1,\\
  &z\in H_{\mathrm{crit}}\}.
  \end{aligned}
  \label{eq:supp_height_set}
\end{equation}
The static geometric risk is then obtained from a monotonic height-to-risk mapping:
\begin{equation}
  R_{\mathrm{geo}}(x,y)=
  \begin{cases}
  \psi_h\!\left(\max \mathcal{Z}_{x,y}\right), & \mathcal{Z}_{x,y}\neq \emptyset,\\
  0, & \mathcal{Z}_{x,y}= \emptyset,
  \end{cases}
  \label{eq:supp_height_risk}
\end{equation}
where $\psi_h(\cdot)$ maps collision-relevant occupied height to the global static risk range $[0.2,0.7]$. Semantic priors are incorporated by an additional category-dependent term $\psi_s(c)$, especially for vulnerable road users. We first define
\begin{equation}
  \eta(x,y)=R_{\mathrm{geo}}(x,y)+\psi_s(c_{x,y}).
  \label{eq:supp_static_score}
\end{equation}
The static risk is assigned only to cells with collision-relevant occupancy:
\begin{equation}
  R_{\mathrm{static}}(x,y)
  =
  \begin{cases}
  \mathrm{clip}\bigl(\eta(x,y),\,0.2,\,0.7\bigr),
  & \mathcal{Z}_{x,y}\neq \emptyset,\\
  0, & \mathcal{Z}_{x,y}= \emptyset.
  \end{cases}
  \label{eq:supp_static_risk}
\end{equation}

Map-derived regulatory risk is represented separately:
\begin{equation}
  R_{\mathrm{reg}}(x,y)=
  \begin{cases}
  0.1, & (x,y)\in \mathcal{M}_{\mathrm{reg}},\\
  0, & \mathrm{otherwise},
  \end{cases}
  \label{eq:supp_reg_risk}
\end{equation}
where $\mathcal{M}_{\mathrm{reg}}$ contains lane dividers, stop lines, crosswalks, and related road markings. For unobservable but relevant cells, a small positive uncertainty prior is used:
\begin{equation}
  R_{\mathrm{unc}}(x,y)=0.05\cdot \mathbf{1}\bigl[(x,y)\in\mathcal{U}\bigr],
  \label{eq:supp_unc_risk}
\end{equation}
where $\mathcal{U}$ denotes blind-spot or undefined cells. The global risk-aware occupancy map is the maximum over these risk sources:
\begin{equation}
  R_{\mathrm{global}}(x,y)
  = \max\bigl(
  R_{\mathrm{static}}(x,y),
  R_{\mathrm{reg}}(x,y),
  R_{\mathrm{unc}}(x,y)
  \bigr).
  \label{eq:supp_global_risk}
\end{equation}

Dynamic risk is stored as a separate branch to avoid mixing static and temporal supervision. Let $\mathcal{A}_{\mathrm{dyn}}$ be the set of agents with velocity above $0.5\,\mathrm{m/s}$, and let $K=5$ denote the future steps over $2.5$\,s with a $0.5$\,s interval. If $B_{a,k}$ is the BEV footprint of dynamic agent $a$ at future step $k$, then the initial dynamic risk can be written as:
\begin{equation}
  \begin{aligned}
  \hat{R}_{\mathrm{dyn}}(x,y)
  = \max_{\substack{a\in\mathcal{A}_{\mathrm{dyn}}\\ k\in\{1,\dots,K\}}}
  &\mathbf{1}\bigl[(x,y)\in B_{a,k}\bigr]\\
  &\cdot\left(0.5+0.5\frac{K-k}{K-1}\right).
  \end{aligned}
  \label{eq:supp_dynamic_initial}
\end{equation}
This assigns larger risk to nearer future states and lower risk to later future states. A Gaussian diffusion kernel then converts the discrete future footprints into a continuous risk field:
\begin{equation}
  R_{\mathrm{dyn}} = \mathcal{G}_{\sigma} * \hat{R}_{\mathrm{dyn}},
  \label{eq:supp_dynamic_diffusion}
\end{equation}
where $*$ denotes spatial convolution. The planner can query $R_{\mathrm{global}}$ and $R_{\mathrm{dyn}}$ separately, or use their maximum when a single risk score is needed:
\begin{equation}
  R_{\mathrm{plan}}(x,y)=
  \max\bigl(R_{\mathrm{global}}(x,y),R_{\mathrm{dyn}}(x,y)\bigr).
  \label{eq:supp_planning_risk}
\end{equation}

\section{B. RiskOcc Map Construction Details}

The four annotation pipelines in the main paper can be summarized in a unified notation. The static pipeline first collapses 3D occupancy into BEV through the height-restricted projection in Eq.~\eqref{eq:supp_height_set}. To reduce quantization artifacts, the raw map $R_{\mathrm{init}}$ is refined through morphological closing and local infilling:
\begin{equation}
  \hat{R}_{0}
  =\mathcal{F}\!\left((R_{\mathrm{init}}\oplus K_d)\ominus K_e\right),
  \label{eq:supp_refine}
\end{equation}
where $\oplus$ and $\ominus$ denote dilation and erosion, $K_d$ and $K_e$ are structuring elements, and $\mathcal{F}(\cdot)$ denotes local context propagation for sparse-region repair.

The lane-line and drivable-area pipelines are fused with the static map through deterministic maximum aggregation:
\begin{equation}
  R_{\mathrm{global}}
  = \max\bigl(
  \hat{R}_{0},
  R_{\mathrm{reg}},
  R_{\mathrm{drive}},
  R_{\mathrm{unc}}
  \bigr).
  \label{eq:supp_pipeline_fusion}
\end{equation}
Here $R_{\mathrm{drive}}$ is the drivable-road prior from the HD map, and $R_{\mathrm{unc}}$ is the low uncertainty prior in Eq.~\eqref{eq:supp_unc_risk}. Dynamic annotations are produced with Eq.~\eqref{eq:supp_dynamic_initial} and Eq.~\eqref{eq:supp_dynamic_diffusion}. This construction preserves regulatory constraints and future interaction risk without requiring a full 4D voxel tensor.

\section{C. ROIDrive Fusion Details}

\subsection{SDMap+ Prior Fusion}

SDMap+ is used as a lightweight map prior to compensate for static scene elements that may be difficult to infer from onboard cameras alone. Given the ego pose, a local map crop is obtained from SDMap+ and converted into a risk-aware occupancy prior through a map-element-to-risk lookup table. The predicted global risk field and the SDMap+ prior are fused by cell-wise maximum:
\begin{equation}
    R_{\mathrm{fused}}(x,y)=
    \max\bigl(R_{\mathrm{pred}}(x,y), R_{\mathrm{map}}(x,y)\bigr).
    \label{eq:supp_sdmap_fusion}
\end{equation}
The max operator preserves the stronger risk evidence from either source, allowing the network prediction to handle dynamic and visually observed hazards while the prior map fills static regulatory or drivable-space constraints.

\begin{figure*}[t]
    \centering
    \includegraphics[width=0.99\textwidth]{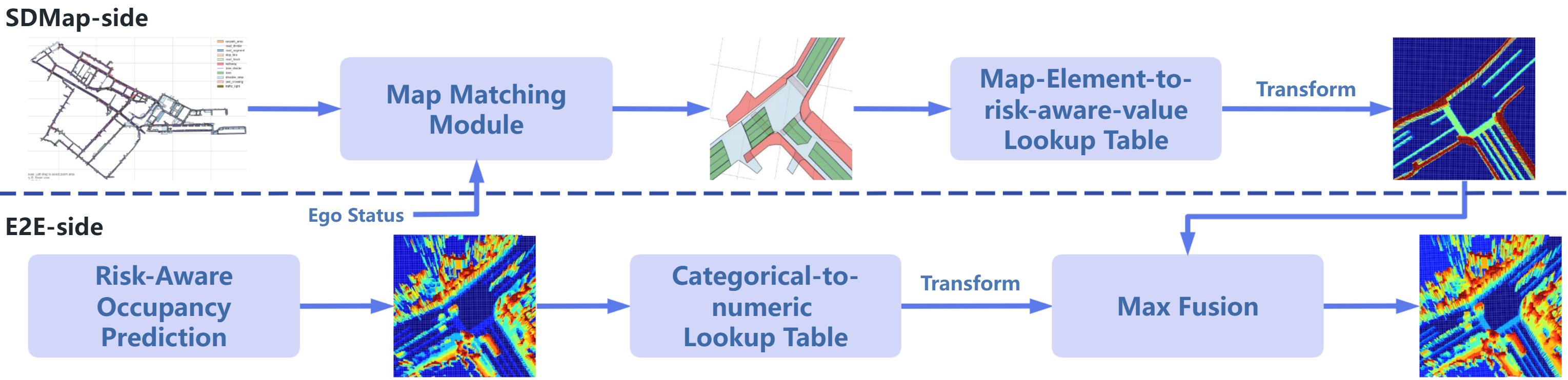}
    \caption{SDMap+ embedding and fusion module. On the cloud side, map matching locates the ego vehicle in the prior urban map and extracts nearby map elements. These elements are transformed into a risk-aware occupancy prior through a map-element-to-risk lookup table. On the vehicle side, the predicted RiskOcc map is converted through a category-to-numeric lookup table. The two risk maps are fused by cell-wise maximum to obtain an enhanced global risk field.}
    \label{fig:supp_sdmap}
\end{figure*}

\subsection{Sliding-Window Risk Tokenization}

The fused risk-aware occupancy map is converted into local risk tokens before instance fusion. Sliding-window sampling extracts 256 local regions, each with a spatial size of $40\times40$. Each local patch is flattened or projected into a feature vector, concatenated with its absolute positional embedding, and transformed by an MLP:
\begin{equation}
    \mathrm{LRM}=\mathrm{MLP}\bigl([\mathrm{SWS}(\mathrm{ROM}); \mathrm{PE}]\bigr),
    \label{eq:supp_sws}
\end{equation}
where $\mathrm{ROM}$ is the risk-aware occupancy map, $\mathrm{SWS}(\cdot)$ denotes sliding-window sampling, $\mathrm{PE}$ is the positional embedding, and $\mathrm{LRM}$ is the set of local risk-aware occupancy tokens. These tokens are then fused with instance tokens through cross-attention, enabling each ego or agent token to attend to nearby dense risk evidence.

\begin{figure}[t]
    \centering
    \includegraphics[width=0.47\textwidth]{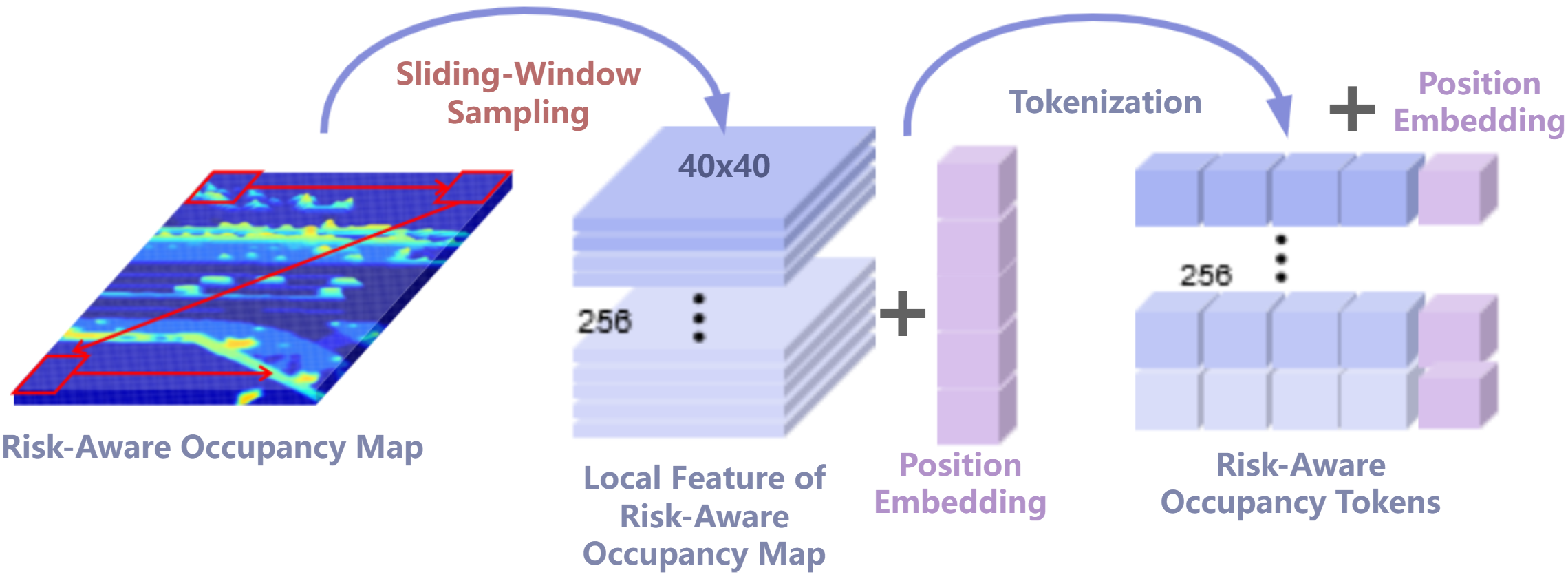}
    \caption{Sliding-window tokenization of the risk-aware occupancy map. The dense risk map is decomposed into 256 local $40\times40$ regions. Each local risk feature is combined with a positional embedding and projected into risk-aware occupancy tokens that match the dimensionality of instance tokens.}
    \label{fig:supp_sliding_window}
\end{figure}

\subsection{Temporal Query Consistency}

The main paper describes TQC as an inference-time query reliability gate. Here we expand its computation. Let $p_i^t$ and $q_i^t$ denote the position and feature of query $i$ at time $t$. Previous-frame positions are first transformed into the current ego frame:
\begin{equation}
  \tilde{p}_j^{t-1}=T_{t-1\rightarrow t}\,p_j^{t-1},
  \label{eq:supp_ego_align}
\end{equation}
where $T_{t-1\rightarrow t}$ is computed from ego-motion. Each current query is matched to its nearest previous query:
\begin{equation}
  m(i)=\arg\min_j \|p_i^t-\tilde{p}_j^{t-1}\|_2.
  \label{eq:supp_query_match}
\end{equation}
The feature and spatial consistency terms are:
\begin{equation}
  \begin{aligned}
  s_i^{\mathrm{feat}}
  &=\frac{1+\cos(q_i^t,q_{m(i)}^{t-1})}{2},\\
  s_i^{\mathrm{pos}}
  &=\exp\!\left(
  -\frac{\|p_i^t-\tilde{p}_{m(i)}^{t-1}\|_2^2}
  {2\sigma_p^2}
  \right).
  \end{aligned}
  \label{eq:supp_consistency_terms}
\end{equation}
The query reliability weight is:
\begin{equation}
  \begin{aligned}
  w_i=\max\Bigl(
  w_{\min},
  (s_i^{\mathrm{feat}})^{\alpha}
  (s_i^{\mathrm{pos}})^{1-\alpha}
  \Bigr).
  \end{aligned}
  \label{eq:supp_query_weight}
\end{equation}
For newly appearing unmatched queries, $w_i$ is set to a fixed fallback weight. Pairwise attention reliability is computed by a geometric mean:
\begin{equation}
  g_{ij}=\sqrt{w_iw_j}.
  \label{eq:supp_pair_gate}
\end{equation}
The gate is converted to an attention logit bias:
\begin{equation}
  \beta_{ij}
  =\log\frac{g_{ij}+\epsilon}{1-g_{ij}+\epsilon}.
  \label{eq:supp_logit_bias}
\end{equation}
The resulting motion self-attention is:
\begin{equation}
  \mathrm{Attn}(Q,K,V)
  =\mathrm{softmax}\!\left(
  \frac{QK^{\top}}{\sqrt{d}}+\mathrm{sg}(\beta)
  \right)V,
  \label{eq:supp_tqc_attention}
\end{equation}
where $\mathrm{sg}(\cdot)$ denotes stop-gradient. This design suppresses flickering or unreliable queries while preserving the base model behavior on the first frame or for newly appearing objects.

\section{D. Training Objective and Evaluation Protocol}

Experiments are conducted on nuScenes, which contains 1000 scenes. We follow the standard split with 700 scenes for training, 150 for validation, and 150 for testing. Open-loop planning is evaluated at 1\,s, 2\,s, and 3\,s horizons using L2 displacement error and collision rate under the UniAD and ST-P3 protocols.

\subsection{Training Losses}

The training objective combines perception and planning losses. The perception loss includes object detection, map perception, and risk-aware occupancy losses. The RiskOcc branch uses classification and regression losses, including Dice loss, Lovasz-Softmax loss, SmoothL1 loss, semantic-scene-affinity loss, and SSIM loss. The Dice loss is:
\begin{equation}
  \begin{split}
    \mathcal{L}_{\mathrm{dice}} =
    1 - \frac{1}{C}\sum_{c=1}^{C}
    \frac{2\sum_{i=1}^{N}Y_{i,c}\hat{Y}_{i,c}+\epsilon}
    {\sum_{i=1}^{N}Y_{i,c}^{2}
    +\sum_{i=1}^{N}\hat{Y}_{i,c}^{2}+\epsilon}.
  \end{split}
    \label{eq:supp_dice}
\end{equation}
where $C$ is the number of risk categories, $N$ is the number of samples, $Y_{i,c}$ is the ground-truth label, $\hat{Y}_{i,c}$ is the predicted probability, and $\epsilon$ is a smoothing constant. The SmoothL1 loss for continuous risk regression is:
\begin{equation}
\mathcal{L}_{\mathrm{smooth}\text{-}L1} =
\begin{cases}
\frac{1}{2\beta}(y-\hat{y})^2, & |y-\hat{y}|\leq \beta,\\
|y-\hat{y}|-\frac{\beta}{2}, & |y-\hat{y}|>\beta.
\end{cases}
\label{eq:supp_smooth_l1}
\end{equation}

The planning objective follows the generative trajectory planning formulation and includes trajectory regression, interaction constraints, trajectory classification, and latent-space regularization:
\begin{equation}
  \begin{split}
    \mathcal{L}_{\mathrm{plan}} =
    L_{\mathrm{tra}}(\hat{\mathbf{T}}_e,\mathbf{T}_e)
    + \frac{1}{N_a}L_{\mathrm{tra}}(\hat{\mathbf{T}}_a,\mathbf{T}_a)
    \\ + \lambda_c L_{\mathrm{focal}}
    (\hat{\mathbf{C}}_a,\mathbf{C}_a).
  \end{split}
    \label{eq:supp_plan_loss}
\end{equation}
where $L_{\mathrm{tra}}$ includes L1 trajectory discrepancy and planning constraints such as ego-agent collision, ego-boundary overstepping, and lane-direction consistency. The latent distribution is regularized by:
\begin{equation}
    \mathcal{L}_{\mathrm{KL}} =
    D_{\mathrm{KL}}\left(p(\mathbf{z}\mid\mathbf{I}),\,
    p(\mathbf{z}\mid\mathbf{T})\right),
    \label{eq:supp_kl_loss}
\end{equation}
where $\mathbf{I}$ denotes the input observations and $\mathbf{T}$ denotes trajectory supervision.

\begin{table*}[t!]
\centering

\fontsize{7.5}{8.5}\selectfont
\setlength{\tabcolsep}{0pt}
\renewcommand{\arraystretch}{1.15}
\begin{tabular*}{\textwidth}{@{\extracolsep{\fill}}llcccccccccccccccc@{}}

\toprule
\multirow{3}{*}{\textbf{Method}} & \multirow{3}{*}{\textbf{Venue}}
& \multicolumn{4}{c}{\textbf{UniAD L2 $\downarrow$}}
& \multicolumn{4}{c}{\textbf{UniAD Collision $\downarrow$}}
& \multicolumn{4}{c}{\textbf{ST-P3 L2 $\downarrow$}}
& \multicolumn{4}{c}{\textbf{ST-P3 Collision $\downarrow$}}\\
\cmidrule(lr){3-6}\cmidrule(lr){7-10}\cmidrule(lr){11-14}\cmidrule(l){15-18}
& & 1s & 2s & 3s & \cellcolor[HTML]{DCDCDC}Avg.
& 1s & 2s & 3s & \cellcolor[HTML]{DCDCDC}Avg.
& 1s & 2s & 3s & \cellcolor[HTML]{DCDCDC}Avg.
& 1s & 2s & 3s & \cellcolor[HTML]{DCDCDC}Avg.\\
\midrule
UniAD & CVPR 23
& 0.48 & 0.96 & 1.65 & \cellcolor[HTML]{DCDCDC}1.03
& 0.05 & 0.17 & 0.71 & \cellcolor[HTML]{DCDCDC}0.31
& N/A & N/A & N/A & \cellcolor[HTML]{DCDCDC}N/A
& N/A & N/A & N/A & \cellcolor[HTML]{DCDCDC}N/A\\
ST-P3 & ECCV 22
& N/A & N/A & N/A & \cellcolor[HTML]{DCDCDC}N/A
& N/A & N/A & N/A & \cellcolor[HTML]{DCDCDC}N/A
& 1.33 & 2.11 & 2.90 & \cellcolor[HTML]{DCDCDC}2.11
& 0.23 & 0.62 & 1.27 & \cellcolor[HTML]{DCDCDC}0.71\\
OccNet & ICCV 23
& N/A & N/A & N/A & \cellcolor[HTML]{DCDCDC}N/A
& N/A & N/A & N/A & \cellcolor[HTML]{DCDCDC}N/A
& 1.29 & 2.13 & 2.99 & \cellcolor[HTML]{DCDCDC}2.13
& 0.21 & 0.59 & 1.37 & \cellcolor[HTML]{DCDCDC}0.72\\
VAD-Tiny & ICCV 23
& N/A & N/A & N/A & \cellcolor[HTML]{DCDCDC}N/A
& N/A & N/A & N/A & \cellcolor[HTML]{DCDCDC}N/A
& 0.46 & 0.76 & 1.12 & \cellcolor[HTML]{DCDCDC}0.78
& 0.21 & 0.35 & 0.58 & \cellcolor[HTML]{DCDCDC}0.38\\
VAD-Base & ICCV 23
& 0.50 & 1.02 & 1.69 & \cellcolor[HTML]{DCDCDC}1.07
& 0.01 & 0.30 & 0.95 & \cellcolor[HTML]{DCDCDC}0.42
& 0.41 & 0.70 & 1.05 & \cellcolor[HTML]{DCDCDC}0.72
& 0.07 & 0.17 & 0.41 & \cellcolor[HTML]{DCDCDC}0.22\\
OccWorld & ECCV 24
& N/A & N/A & N/A & \cellcolor[HTML]{DCDCDC}N/A
& N/A & N/A & N/A & \cellcolor[HTML]{DCDCDC}N/A
& 0.43 & 1.08 & 1.99 & \cellcolor[HTML]{DCDCDC}1.17
& 0.07 & 0.38 & 1.35 & \cellcolor[HTML]{DCDCDC}0.60\\
PPAD (P.) & ECCV 24
& 0.38 & 0.83 & 1.45 & \cellcolor[HTML]{DCDCDC}0.89
& 0.02 & 0.20 & 0.93 & \cellcolor[HTML]{DCDCDC}0.38
& 0.38 & 0.69 & 1.26 & \cellcolor[HTML]{DCDCDC}0.75
& 0.08 & 0.12 & 0.38 & \cellcolor[HTML]{DCDCDC}0.19\\
PARADrive & CVPR 24
& 0.40 & 0.77 & 1.31 & \cellcolor[HTML]{DCDCDC}0.83
& 0.07 & 0.25 & 0.60 & \cellcolor[HTML]{DCDCDC}0.30
& N/A & N/A & N/A & \cellcolor[HTML]{DCDCDC}N/A
& N/A & N/A & N/A & \cellcolor[HTML]{DCDCDC}N/A\\
BEVPlanner & CVPR 24
& 0.30 & 0.52 & 0.83 & \cellcolor[HTML]{DCDCDC}0.55
& 0.10 & 0.37 & 1.30 & \cellcolor[HTML]{DCDCDC}0.59
& N/A & N/A & N/A & \cellcolor[HTML]{DCDCDC}N/A
& N/A & N/A & N/A & \cellcolor[HTML]{DCDCDC}N/A\\
Epona & ICCV 25
& N/A & N/A & N/A & \cellcolor[HTML]{DCDCDC}N/A
& N/A & N/A & N/A & \cellcolor[HTML]{DCDCDC}N/A
& 0.61 & 1.17 & 1.98 & \cellcolor[HTML]{DCDCDC}1.25
& 0.01 & 0.22 & 0.85 & \cellcolor[HTML]{DCDCDC}0.36\\
LAW* & ICLR 25
& 0.26 & 0.57 & 1.01 & \cellcolor[HTML]{DCDCDC}0.61
& 0.14 & 0.21 & 0.54 & \cellcolor[HTML]{DCDCDC}0.30
& N/A & N/A & N/A & \cellcolor[HTML]{DCDCDC}N/A
& N/A & N/A & N/A & \cellcolor[HTML]{DCDCDC}N/A\\
LAW & ICLR 25
& 0.24 & \textbf{0.46} & \textbf{0.76} & \cellcolor[HTML]{DCDCDC}\textbf{0.49}
& 0.08 & 0.10 & 0.39 & \cellcolor[HTML]{DCDCDC}0.19
& N/A & N/A & N/A & \cellcolor[HTML]{DCDCDC}N/A
& N/A & N/A & N/A & \cellcolor[HTML]{DCDCDC}N/A\\
PreWorld & ICLR 25
& 0.49 & 1.22 & 2.32 & \cellcolor[HTML]{DCDCDC}1.34
& 0.19 & 0.57 & 2.65 & \cellcolor[HTML]{DCDCDC}1.14
& N/A & N/A & N/A & \cellcolor[HTML]{DCDCDC}N/A
& N/A & N/A & N/A & \cellcolor[HTML]{DCDCDC}N/A\\
ELM & ECCV 24
& 0.34 & 1.23 & 2.57 & \cellcolor[HTML]{DCDCDC}1.38
& 0.12 & 0.50 & 2.36 & \cellcolor[HTML]{DCDCDC}0.99
& N/A & N/A & N/A & \cellcolor[HTML]{DCDCDC}N/A
& N/A & N/A & N/A & \cellcolor[HTML]{DCDCDC}N/A\\
FeD & CVPR 24
& 0.27 & 0.53 & 0.94 & \cellcolor[HTML]{DCDCDC}0.58
& \textbf{0.00} & \textbf{0.04} & 0.52 & \cellcolor[HTML]{DCDCDC}0.19
& N/A & N/A & N/A & \cellcolor[HTML]{DCDCDC}N/A
& N/A & N/A & N/A & \cellcolor[HTML]{DCDCDC}N/A\\
World4Drive & ICCV 25
& 0.23 & 0.47 & 0.81 & \cellcolor[HTML]{DCDCDC}0.50
& 0.02 & 0.12 & \textbf{0.33} & \cellcolor[HTML]{DCDCDC}\textbf{0.16}
& N/A & N/A & N/A & \cellcolor[HTML]{DCDCDC}N/A
& N/A & N/A & N/A & \cellcolor[HTML]{DCDCDC}N/A\\
GenAD & ECCV 24
& 0.33 & 0.81 & 1.58 & \cellcolor[HTML]{DCDCDC}0.91
& 0.05 & 0.29 & 0.87 & \cellcolor[HTML]{DCDCDC}0.40
& 0.25 & 0.46 & 0.76 & \cellcolor[HTML]{DCDCDC}0.49
& 0.06 & 0.15 & 0.32 & \cellcolor[HTML]{DCDCDC}0.17\\
\rowcolor[HTML]{F0F0F0}\textbf{ROIDrive} & N/A
& \textbf{0.21} & 0.59 & 1.26 & \cellcolor[HTML]{DCDCDC}0.69
& \textbf{0.00} & 0.14 & 0.86 & \cellcolor[HTML]{DCDCDC}0.33
& \textbf{0.16} & \textbf{0.32} & \textbf{0.57} & \cellcolor[HTML]{DCDCDC}\textbf{0.35}
& \textbf{0.00} & \textbf{0.04} & \textbf{0.21} & \cellcolor[HTML]{DCDCDC}\textbf{0.08}\\
\bottomrule
\end{tabular*}
\caption{Complete open-loop planning performance on the nuScenes validation set. L2 denotes trajectory displacement error in meters, and collision rate is reported in percentage. N/A indicates that a metric is unavailable under the corresponding protocol. Best values are shown in \textbf{bold}.}
\label{tab:supp_full_planning}
\end{table*}

\subsection{Evaluation Metrics}

For risk category prediction, mean intersection-over-union is:
\begin{equation}
  \mathrm{mIoU}
  =\frac{1}{C}\sum_{c=1}^{C}
  \frac{\mathrm{TP}_c}{\mathrm{TP}_c+\mathrm{FP}_c+\mathrm{FN}_c}.
  \label{eq:supp_miou}
\end{equation}
For continuous risk regression, mean absolute error is:
\begin{equation}
  \mathrm{MAE}
  =\frac{1}{|\Omega|}\sum_{(x,y)\in\Omega}
  |\hat{R}(x,y)-R(x,y)|.
  \label{eq:supp_mae}
\end{equation}
MAE@15 and MAE@30 use the same formula over ego-centric $15\,\mathrm{m}\times15\,\mathrm{m}$ and $30\,\mathrm{m}\times30\,\mathrm{m}$ regions. Structural similarity is:
\begin{equation}
  \begin{split}
  \mathrm{SSIM}(X,Y)
  =\frac{(2\mu_X\mu_Y+c_1)(2\sigma_{XY}+c_2)}
  {(\mu_X^2+\mu_Y^2+c_1)(\sigma_X^2+\sigma_Y^2+c_2)}.
  \end{split}
  \label{eq:supp_ssim}
\end{equation}
Dynamic risk prediction is evaluated by Mask-MAE to avoid domination by background zeros:
\begin{equation}
  \begin{aligned}
  M_i=\mathbf{1}[\hat{R}_i>\tau\;\mathrm{or}\;R_i>\tau],
  \\
  \mathrm{Mask\mbox{-}MAE}
  =\frac{1}{\sum_i M_i}\sum_i M_i|\hat{R}_i-R_i|.
  \end{aligned}
  \label{eq:supp_mask_mae}
\end{equation}
For planning, the horizon-specific L2 error is:
\begin{equation}
  \mathrm{L2}(t)=\|\hat{T}_e(t)-T_e(t)\|_2,
  \label{eq:supp_l2}
\end{equation}
and collision rate is the percentage of evaluated samples in which the planned ego trajectory overlaps with another traffic participant under the corresponding benchmark protocol.

\section{E. Full Open-Loop Planning Results}

Table~\ref{tab:supp_full_planning} reports the complete open-loop planning comparison on the nuScenes validation set under both protocols. It complements the collision-focused main-paper table by reporting the complete ROIDrive model with all modules enabled. Compared with GenAD, ROIDrive reduces average L2 from 0.91\,m to 0.69\,m under UniAD and from 0.49\,m to 0.35\,m under ST-P3. Average collision rate decreases from 0.40\% to 0.33\% under UniAD and from 0.17\% to 0.08\% under ST-P3.

\section{F. Qualitative Scene Visualizations}

\begin{figure*}[p]
\centering

\begin{subfigure}{\textwidth}
\centering
\maybeincludegraphics[width=0.86\linewidth, trim=5 5 5 5, clip]{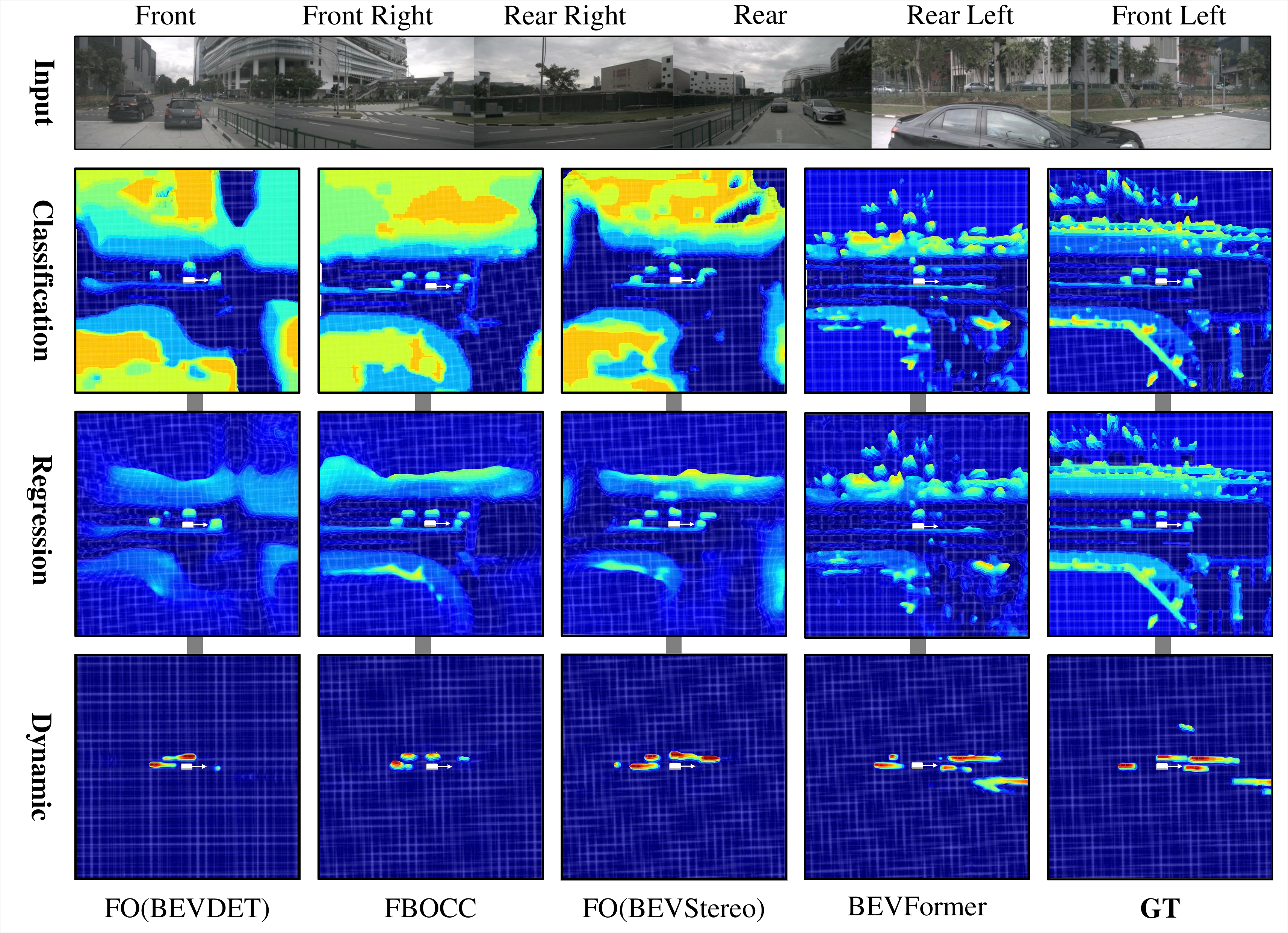}
\caption{Typical urban driving scenario.}
\label{fig:visualization_1}
\end{subfigure}

\vspace{0.5em}

\begin{subfigure}{\textwidth}
\centering
\maybeincludegraphics[width=0.86\linewidth, trim=5 5 5 5, clip]{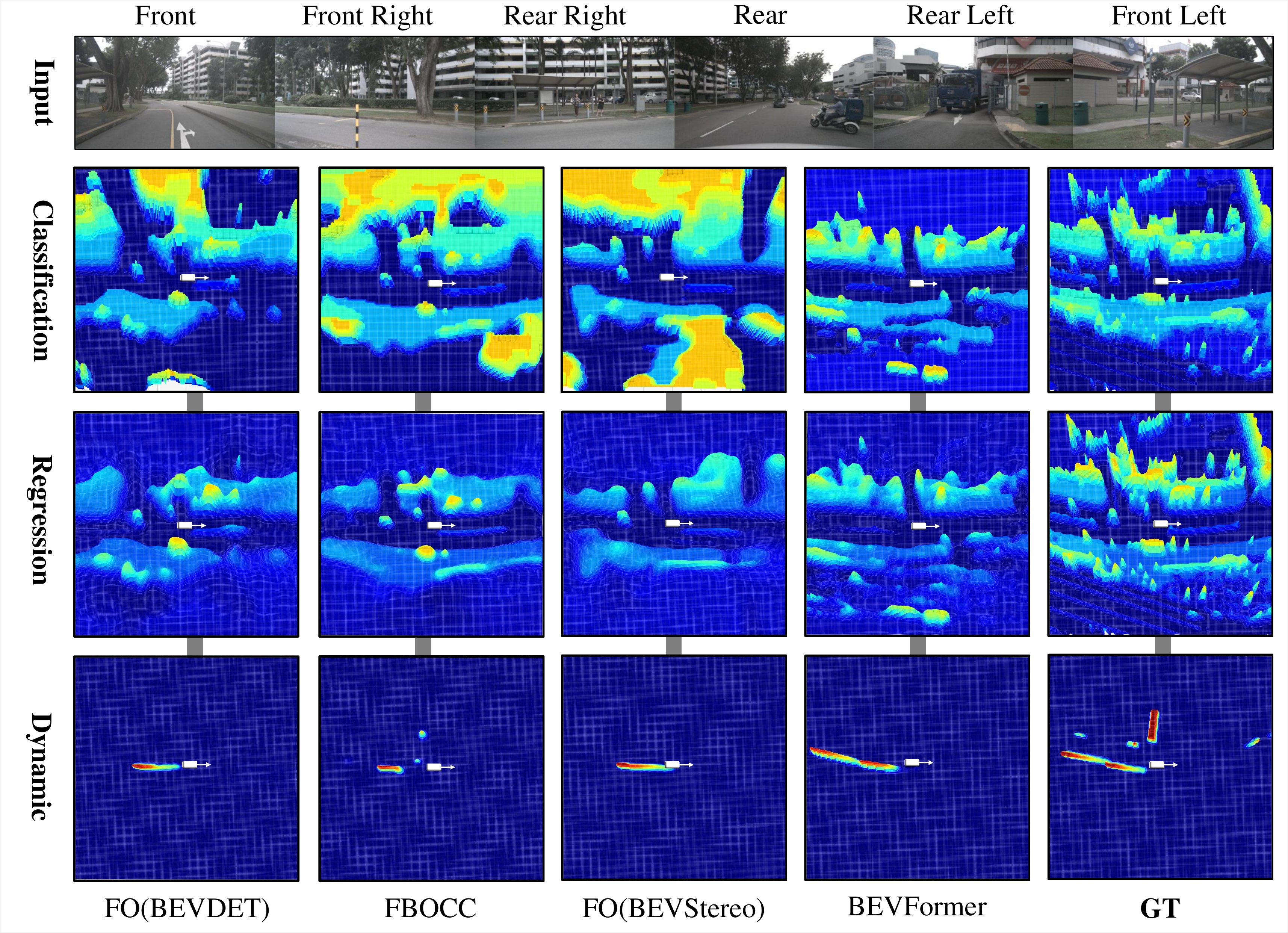}
\caption{Dense urban interaction scenario.}
\label{fig:visualization_4}
\end{subfigure}

\caption{Qualitative visualization of RiskOcc4D-nuScenes predictions in urban scenes. Classification maps expose coarse spatial coverage, regression maps show static risk continuity, and dynamic maps indicate whether moving-agent risk is localized near future occupied regions.}
\label{fig:visualization_urban}
\end{figure*}

\begin{figure*}[p]
\centering

\begin{subfigure}{\textwidth}
\centering
\maybeincludegraphics[width=0.86\linewidth, trim=5 5 5 5, clip]{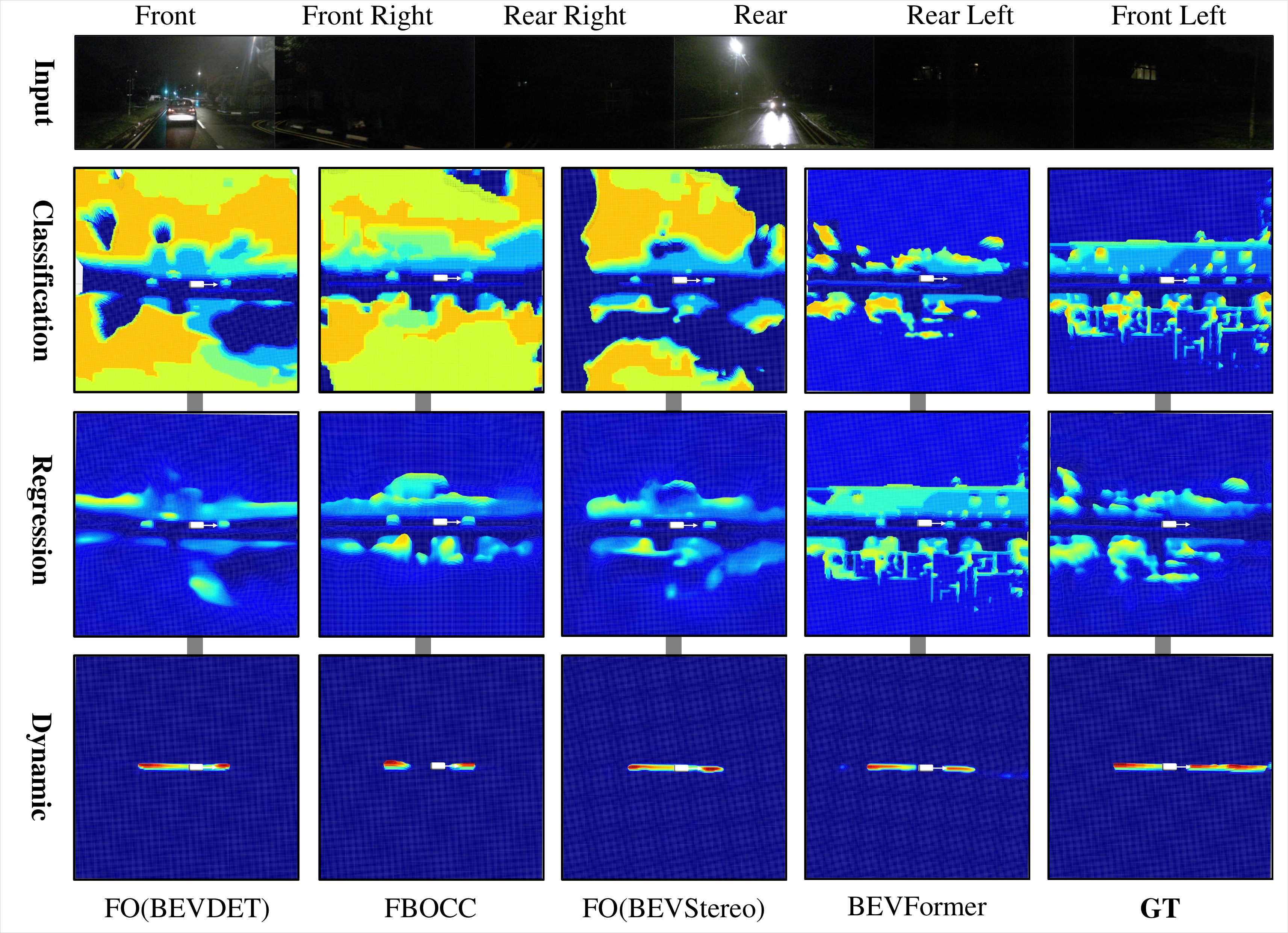}
\caption{Nighttime driving scenario.}
\label{fig:visualization_3}
\end{subfigure}

\vspace{0.5em}

\begin{subfigure}{\textwidth}
\centering
\maybeincludegraphics[width=0.86\linewidth, trim=5 5 5 5, clip]{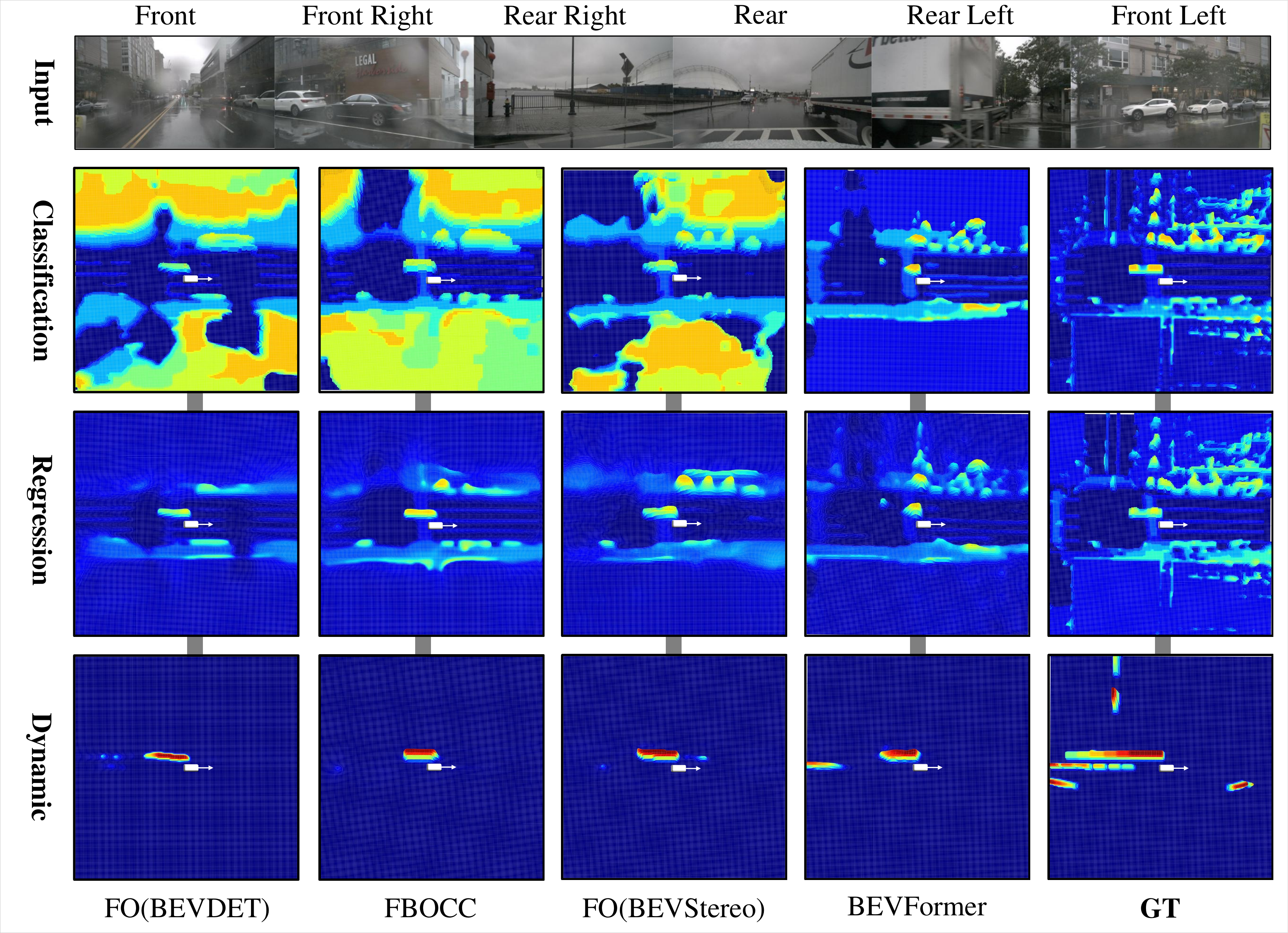}
\caption{Rainy-weather driving scenario.}
\label{fig:visualization_2}
\end{subfigure}

\caption{Qualitative visualization of RiskOcc4D-nuScenes predictions under degraded visibility. Nighttime and rainy-weather scenes reveal failure modes such as over-smoothed static risk, false positive background activation, and missed dynamic regions.}
\label{fig:visualization_adverse}
\end{figure*}

The qualitative examples in Figures~\ref{fig:visualization_urban} and~\ref{fig:visualization_adverse} complement the quantitative results by showing how risk-aware occupancy behaves across different driving conditions. Each panel contains six surround-view input images, followed by three predicted risk maps: discrete classification, continuous regression, and dynamic future risk. Columns compare representative occupancy backbones, including FO(BEVDET), FBOCC, FO(BEVStereo), and BEVFormer, with the ground truth shown in the rightmost column. These visualizations inspect whether a method preserves road geometry, suppresses background artifacts, and localizes dynamic risk in a planning-relevant BEV layout.

\noindent\textbf{Typical urban driving scenario.}
This daytime scene has clear surround-view evidence, visible vehicles ahead, and regular urban road boundaries. It mainly examines whether the predicted RiskOcc maps recover the broad drivable/non-drivable layout while keeping static boundaries continuous. The classification maps expose coarse spatial coverage, whereas the regression maps should smooth grid artifacts without erasing road-edge structure. In the dynamic row, a desirable prediction keeps compact high-risk responses around the nearby moving agents rather than activating free space.

\noindent\textbf{Dense urban interaction scenario.}
This scene contains denser roadside structures, multiple traffic participants, and stronger foreground-background clutter. It stresses whether the occupancy backbones can separate true static risk from visually complex urban context. Compared with the typical case, the ground truth contains richer boundary structure and a more complex dynamic pattern. The key observation is not only whether occupied regions are detected, but also whether future dynamic risk remains localized and complete instead of being fragmented, missed, or spread into neighboring lanes.

\noindent\textbf{Nighttime driving scenario.}
The nighttime example tests robustness under weak illumination, dark side views, and headlight glare. Because image texture is limited, RiskOcc prediction should rely on consistent BEV structure rather than amplifying ambiguous dark regions. The classification maps may become blocky or overconfident under low visibility, so the regression maps are expected to provide smoother static-risk continuity. In the dynamic row, the main requirement is to retain the compact risk of visible leading agents while suppressing background activation.

\noindent\textbf{Rainy-weather driving scenario.}
The rainy-weather example further evaluates robustness under blurred cameras, low contrast, and water-induced visual artifacts. Reliable predictions should preserve the road layout and nearby obstacle structure without converting rain streaks or reflections into large false high-risk regions. The dynamic maps are especially important here because adverse weather can make moving objects less distinct; the predicted future-risk responses should remain aligned with traffic participants and should not collapse into either missing detections or diffuse background noise.

Together, these qualitative checks complement mIoU, MAE, SSIM, and Mask-MAE because they expose spatial failure modes that may be averaged out in aggregate metrics.